\documentclass[journal]{IEEEtran}

\usepackage{times}
\usepackage{epsfig}
\usepackage{graphicx}
\usepackage{amsmath}
\usepackage{amssymb}
\usepackage{algorithm}
\usepackage{algorithmic}
\usepackage[sort&compress,numbers]{natbib}
\usepackage{multirow}
\usepackage{makecell}
\usepackage[switch]{lineno}

\usepackage[pagebackref=false,colorlinks=true,urlcolor = blue,bookmarks=false,]{hyperref}
\begin{document}
	%
	\title{RGB-T Object Tracking: Benchmark and Baseline}
	%
	%
	%
	\author{Chenglong~Li,
		Xinyan~Liang,
		Yijuan~Lu,
		Nan~Zhao,
		and~Jin~Tang
		\thanks{C. Li, X. Liang, N. Zhao and J. Tang are with the School of Computer Science and Technology, Anhui University, Hefei 230601, China. (lcl1314@foxmail.com, lxy2626@foxmail.com, zhn1528@gmail.com, jtang99029@foxmail.com)}
		\thanks{Y. Lu is with the department of computer science, Texas State University, San Marcos, USA. (lu@txstate.edu)}}

	\markboth{Submission to IEEE Transactions on Image Processing}%
	{Shell \MakeLowercase{\textit{et al.}}: Bare Demo of IEEEtran.cls for IEEE Journals}
	%



    \maketitle
	
	\begin{abstract}
		RGB-Thermal (RGB-T) object tracking receives more and more attention due to the strongly complementary benefits of thermal information to visible data.
		However, RGB-T research is limited by lacking a comprehensive evaluation platform. In this paper, we propose a large-scale video benchmark dataset for RGB-T tracking.
		 It has three major advantages over existing ones: 1) Its size is sufficiently large for large-scale performance evaluation (total frame number: 234K, maximum frame per sequence: 8K). 2) The alignment between RGB-T sequence pairs is highly accurate, which does not need pre- or post-processing. 3) The occlusion levels are annotated for occlusion-sensitive performance analysis of different tracking algorithms.
		Moreover, we propose a novel graph-based approach to learn a robust object
		representation for RGB-T tracking. In particular, the tracked object is represented with a graph with image patches as nodes. This graph including graph structure, node weights and edge weights is dynamically learned in a unified ADMM (alternating direction method of multipliers)-based optimization framework, in which the modality weights are also incorporated for adaptive fusion of multiple source data.
		Extensive experiments on the large-scale dataset are executed to demonstrate the effectiveness of the proposed tracker against other state-of-the-art tracking methods. We also provide new insights and potential research directions to the field of RGB-T object tracking.
	\end{abstract}
	
	\begin{IEEEkeywords}
		RGB-T tracking, Benchmark dataset, Sparse representation, Dynamic graph learning, Adaptive fusion
	\end{IEEEkeywords}

	%
	\IEEEpeerreviewmaketitle

	\section{Introduction}
	
	\IEEEPARstart{V}{isual} tracking is to estimate states of the target object in subsequent frames, given the initial ground-truth bounding box. It is an active and challenging computer vision task, and has drawn a lot of attentions due to its wide applications, such as video surveillance, self-driving cars, and robotics. Despite of many recent breakthroughs in visual tracking, it still faces many challenging problems especially tracking target objects in various environmental conditions (\emph{e.g.}, low illumination, rain, haze and smog, etc.), which significantly affect the imaging quality of visible spectrum.
	
\begin{figure*}[!t]
		\centering
		\includegraphics[width=2.0\columnwidth]{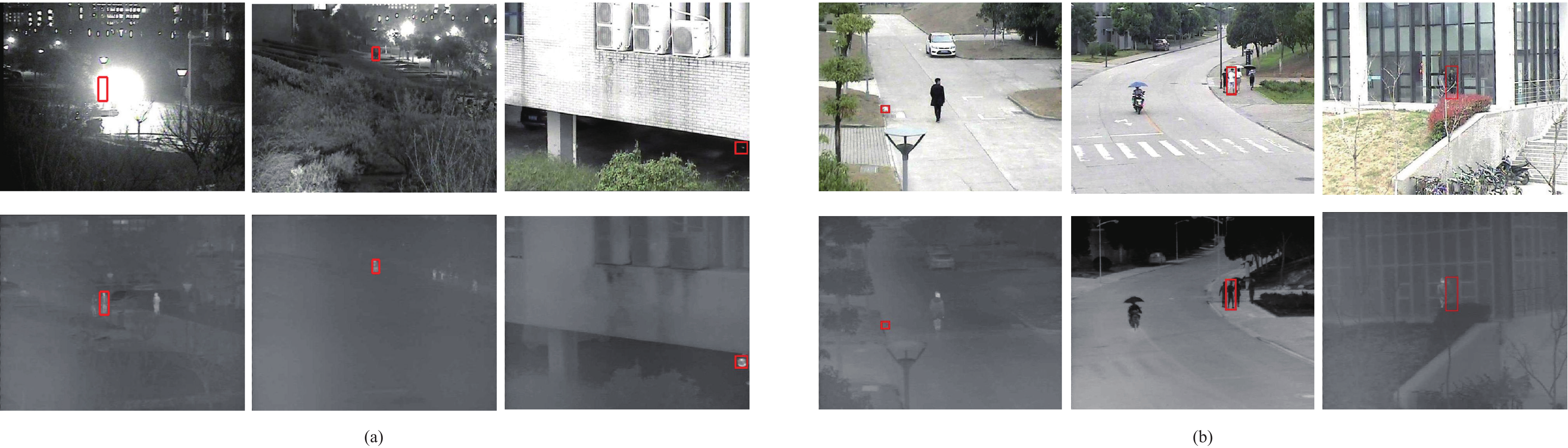}	
		\caption{Illustration of complementary benefits of RGB and thermal images. The first row shows RGB images and the second row shows thermal images. (a) Complementary benefits of thermal images over RGB ones, where visible spectrum is disturbed by high illumination, low illumination while thermal information can overcome them effectively. (b) complementary benefits of RGB images over thermal ones, where thermal spectrum is disturbed by thermal crossover and glass while RGB information can handle them effectively.}
		\label{fig:RGB_T}       
	\end{figure*}

	Integrating visible and thermal (called RGB-T in this paper) spectrum data has been proven to be effective in boosting tracking performance, and also allows tracking target objects in day and night~\cite{ThermalApplications14mva}. Visible and thermal information complement each other and contribute to visual tracking in different aspects. On one hand, thermal infrared camera can capture infrared radiation (0.75-13$\mu m$) emitted by subjects with a temperature above absolute zero.  Thus they are insensitive to lighting conditions and have a strong ability to penetrate haze and smog. These kind of sensors, therefore, are more effective in capturing objects than visible spectrum cameras under poor lighting conditions and bad weathers. On the other hand, visible spectrum cameras are more effective in separating two moving subjects, which are crossing or moving side (called thermal crossover~\cite{Liu12infosci,WELD16tcsvt}). Figure~\ref{fig:RGB_T} shows some typical scenarios.

	There are several video benchmark datasets designed for RGB-T tracking purpose such as OSU-CT~\cite{Davis07cviu}, LITIV~\cite{Torabi12cviu} and GTOT~\cite{Li16tip}. A major issue of these datasets is their limited size (see Table~\ref{tab:datasetComparation} for details). To facilitate this research direction, we contribute a more comprehensive video dataset, called RGBT234 in this paper. RGBT234 includes 234 RGB-T videos, each containing a RGB video and a thermal video. Its total frames reach about 234K and the frame of the biggest video pair reaches 8K. Moreover, we annotate each frame with a minimum bounding box covering the target for both modalities, and also take into full consideration of various environmental challenges, such as raining, night, cold and hot days. To analyze the attribute-based performance of different tracking algorithms, we annotate 12 attributes and implement a dozen of baseline trackers, including deep and non-deep ones (\emph{e.g.}, structured SVM, sparse representation and correlation filter based trackers). In addition, we adopt 5 metrics to evaluate RGB-T trackers.

	Compared with other existing ones, RGBT234 has the following major advantages. 1) It includes a large amount of video frames, which enables users to perform large-scale performance evaluations. 2) Its alignment across modalities is highly accurate, and does not require any pre- or post-processing (\emph{e.g.}, stereo matching~\cite{Li16tip,KrotoskyT07} and color correction~\cite{Hwang_2015_CVPR}) with the advanced imaging mechanisms. 3) Its occlusion levels, including no, partial and heavy occlusions, are annotated for occlusion-sensitive evaluation of different algorithms. 4) Its videos can be captured by both static and moving cameras while keeping the alignment accurate due to the flexibility of our imaging platform. 5) More recent trackers, more attribute annotations and more evaluation metrics are considered.

	With this new benchmark dataset, we investigate how to perform robust object tracking under challenging scenarios by adaptively incorporating the information from RGB-T videos, and propose a novel graph-based algorithm to learn RGB-T object features for online tracking.
	
	Specifically, we partition each target bounding box into a set of non-overlapping patches, which are described with RGB and thermal features. However, background information is inevitably included in the bounding box and likely results in model drifting. Thus, the bounding box cannot represent the target object well. We associate a weight to each patch, which reflects its importance to describe the object. We concatenate all patch features within the target bounding box with their corresponding weights to convey structural information of the object while suppressing the pollution of background. The structured SVM algorithm~\cite{Tsochantaridis05jmlr} is then adopted to perform object tracking.

	In general, the patch weights are computed via a semi-supervised learning algorithm, \emph{e.g.}, manifold ranking~\cite{Zhou04nips} and random walk ~\cite{Kim15iccv}. How to learn an optimal affinity matrix is essential. Most of methods construct a fixed-structure graph (\emph{e.g.}, a 8-neighbor graph where each node is only connected with its 8 neighbor nodes), and thus neglect global intrinsic relationship of patches. Li \emph{et al}.~\cite{Li17aaai} propose to learn a dynamic graph with the low-rank and sparse representations as the graph affinity to capture the global subspace structure of patches. Although this kind of affinity is somehow valid, the meaning of which is already not the same as the original definition~\cite{ijcai/Guo15a}.
	
	Motivated by above observations, we propose a novel graph model to adaptively employ RGB and thermal data to weight learning. We take patches as graph nodes, and pursue a joint sparse representation~\cite{Liu12infosci,Lan14cvpr} with patch feature matrix as input. Note that other constraints on the representation matrix, such as low rank~\cite{LiuLYSYM13}, can also be incorporated in our model. We only employ sparse constraints in this paper for an emphasis on efficient performance. To deal with occasional perturbation or malfunction of individual sources, we assign a weight to each modality to represent the reliability, which allows our method to integrate different spectrum data adaptively. Instead of directly using sparse representations, we learn a more meaningful graph affinity with the assumption that patches are more likely located in the same class (the target object or the background) if their sparse representations have a small distance~\cite{ijcai/Guo15a}. Based on the graph affinity, the patch weights are computed when giving some initial weights to patches~\cite{Kim15iccv}.
	
	It is worth noting that we jointly solve modality weights, sparse representations, and the graph (including graph structure, edge weights and node weights) in a single optimization framework by designing an efficient ADMM (Alternation Direction Method of Multipliers) algorithm~\cite{NIPS2011ADM}.

\begin{table*}[t]\scriptsize	
	\caption{\footnotesize Comparison of RGBT234 against other tracking datasets.}
	\centering
	\begin{tabular}{l l c c c c c c c}
		\hline\noalign{\smallskip}
		Type &  & \multicolumn{7}{c}{Properties}\\
		\noalign{\smallskip}\hline\noalign{\smallskip}
		& &\rotatebox{90}{sequence numbers} & \rotatebox{90}{\# total frames} & \rotatebox{90}{max frames per sequence} & \rotatebox{90}{occlusion annotation} & \rotatebox{90}{attribute annotation} & \rotatebox{90}{moving camera} & \rotatebox{90}{publish year}\\\hline
		\multirow{8}{*}{RGB}
		& OTB50\cite{RGBbenchmark13cvpr}&50 & 29.4K & 3.8K & \checkmark & \checkmark & \checkmark & 2013\\
		& OTB100\cite{RGBbenchmark15pami}&100 & 59K & 3.8K & \checkmark & \checkmark & \checkmark & 2015\\
		& VOT2014\cite{vot14}&25 & 10.3K & 1.2K & \checkmark & \checkmark & \checkmark & 2014\\
		& VOT2015\cite{vot15}&60 & 21.8K & 1.5K & \checkmark & \checkmark & \checkmark & 2015\\
		& Temple-Color\cite{liang2015Tcolor}&128 & 55.3K & 3.8K & \checkmark & \checkmark & \checkmark & 2015\\		
		& ALOV++\cite{Smeulders14pami}&315 & 15.1K & 5.9K & \checkmark & \checkmark & \checkmark & 2013\\
		& NUS-PRO\cite{nus_pro2016}&365 & 135.8K & 5K & \checkmark & \checkmark & \checkmark & 2016\\\hline
		\multirow{3}{*}{Thermal}
		& OSU-T\cite{OSUT2005}&10 & 0.2K & - & \checkmark & \checkmark & - & 2005\\
		& ASL-TID\cite{icra14ASL}&9 & 4.3k & - & - & - & - & 2014\\
		& TIV\cite{cvpr14TIV}& 16 & 63K & 5.9K & - & - & - & 2014\\
		& LTIR\cite{ICCV15TIR}&20 & 11.2K & 1.4K & \checkmark & \checkmark & \checkmark & 2015\\\hline
		\multirow{4}{*}{RGB-T}
		& OSU-CT\cite{Davis07cviu}& 6 & 17K & 2K & \checkmark & \checkmark & - & 2007\\
		& LITIV\cite{Torabi12cviu}& 9 & 6.3K & 1.2K & - & - & - & 2012\\
		& GTOT\cite{Li16tip}& 50 & 15.8K & 0.7K & \checkmark & \checkmark & - & 2016\\
		& RGBT210\cite{Li17rgbt210} &210 & 210K & 8k &\checkmark & \checkmark & \checkmark & 2017 \\
		& {\bf RGBT234}& 234 & 233.8K & 8K & \checkmark & \checkmark & \checkmark & -\\
		\noalign{\smallskip}\hline
	\end{tabular}
	\label{tab:datasetComparation}
\end{table*}
	
	This paper makes the following contributions to RGB-T tracking and related applications.
	
	\begin{itemize}
		\item It contributes a comprehensive RGB-T dataset with 234 fully annotated video sequences for large-scale performance evaluation of different tracking algorithms. A dozen of baseline trackers and 5 evaluation metrics are also included in this benchmark. This benchmark will be open to public~\footnote{RGB-T tracking dataset webpage:\\ \href{https://sites.google.com/view/ahutracking001/} {~~~~https://sites.google.com/view/ahutracking001/}.}.
		
		\item It proposes a novel graph-based optimization algorithm to learn RGB-T object feature representations for robust object tracking. In particular, the graph, including graph structure, edge weights and node weights, and the modality weights are dynamically learned in a unified ADMM-based optimization framework.
		
		\item It carries out extensive experiments on the large-scale benchmark dataset. The evaluation results demonstrate the effectiveness of the proposed approach, and we also provide new insights and potential future research directions for RGB-T object tracking.
	\end{itemize}

This paper provides a more complete understanding of initial results~\cite{Li17rgbt210}, with more background, insights, analysis, and evaluation. In particular, our benchmark advances the previous work~\cite{Li17rgbt210} in several aspects. First, we expand our dataset with more videos by taking more challenges into account. For example, we record some new videos in hot days, which bring new challenges to RGB-T tracking. Second, we select more baseline trackers from recent publications, including deep learning based ones, for fair comparison. Third, we adopt more evaluation metrics used in VOT challenges~\cite{vot15} for comprehensive evaluations of different tracking algorithms. Finally, we carry out extensive experiments on our dataset to evaluate different RGB and RGB-T trackers which demonstrate the effectiveness of the proposed approach. We also provide new insights to the field of RGB-T object tracking and related applications.

\section{Related Work}
According to the relevance to our work, we review related works following three research lines, \emph{ i.e.}, RGB-T tracking datasets, RGB-T Object Tracking and RGB Object Tracking.

\subsection{RGB-T Tracking Datasets}
There are several popular RGB-T video datasets. OSU-CT dataset~\cite{Davis07cviu} contains 6 RGB-T video sequence pairs recorded from two different locations with only people moving, which is obviously not sufficient to evaluate tracking algorithms. Other two RGB-T datasets are collected by~\cite{Torabi12cviu,Bilodeau14ipt}. Most of them suffer from limited size, low diversity, and high bias. Li~\emph{et al.}~\cite{Li17rgbt210} construct a large size RGB-T dataset called RGBT210 with 210 RGB-T pairs. They record videos under a moving platform which enriches much diversity of the dataset, and annotate the challenging attributes for comprehensive evaluation. This dataset, however, has two issues: 1) It lacks the videos captured in hot days, and is thus highly biasd to environmental temperature since thermal sensors are sensitive to temperature. 2) Videos of two modalities share same bounding box annotations, which sometimes are unreasonable due to existence of alignment errors.

Recently, more and more large datasets for RGB tracking have been proposed~\cite{vot15,RGBbenchmark13cvpr,RGBbenchmark15pami,liang2015Tcolor, Smeulders14pami, nus_pro2016}. These datasets bring great benefits to visual tracking. Motivated by the observation, we contribute a larger RGB-T tracking dataset with 234 video pairs in total in this paper. This new dataset includes more challenging RGB-T videos, baseline algorithms, attributes and evaluation metrics. Table~\ref{tab:datasetComparation} shows the detailed comparison of these datasets.

\subsection{RGB-T Object Tracking}
RGB-T object tracking receives more and more attentions in computer vision community with the popularity of thermal infrared sensors~\cite{Cvejic07cvpr, Wu11icif,Liu12infosci,Li17tsmcs,Li16tip}.

Cvejic \emph{et al}.~\cite{Cvejic07cvpr} investigate the impact of pixel-level fusion of videos from grayscale-thermal surveillance cameras. Recently, sparse representation has been successfully applied to RGB-T tracking. Wu \emph{et al}.~\cite{Wu11icif} concatenate the image patches from RGB and thermal sources, and then sparsely represent each sample in the target template space. Liu \emph{et al}.~\cite{Liu12infosci} fuse the tracking results using min operation on the sparse representation coefficients calculated on both RGB and thermal modalities. These methods may limit the tracking performance in dealing with occasional perturbation or malfunction of individual sources as available spectrums contribute equally. Li \emph{et al}.~\cite{Li17tsmcs,Li16tip} introduce a modality weight for each source to represent the imaging quality, and combine the sparse representation in Bayesian filtering framework to perform object tracking.

\subsection{RGB Object Tracking}
The methods for RGB object tracking are vast, and a comprehensive review can be found in~\cite{RGBbenchmark15pami,nus_pro2016}. Here, according to the relevance to our work, we only briefly review the RGB trackers from the following three categories.

\subsubsection{Structured SVM based Trackers}
Hare \emph{et al}. first apply the structured output prediction algorithm~\cite{LMM2005Struck} to visual tracking to avoid the need for an intermediate classification step~\cite{Stuck11iccv}. They further integrate several practical considerations to improve tracking accuracy and efficiency~\cite{hare2016struck}. Recently, associating each patch with a weight in the structured SVM framework has been proven to be an effective way for suppressing the background effects in visual tracking~\cite{Kim15iccv,Li17aaai,Li17regle}. In~\cite{Kim15iccv}, a random walk restart algorithm is utilized on the fixed-structure graph with patches as nodes to compute patch weights within target object bounding box. The above fixed-structure graph is constructed via only local cues, neglecting global cues that are important for exploring the intrinsic relationship among patches. In~\cite{Li17aaai}, the low-rank and sparse representation is used to learn a dynamic graph with global considerations. Li \emph{et al}.~\cite{Li17regle} take both advantages of~\cite{Kim15iccv} and~\cite{Li17aaai} to integrate local and global information in graph learning. Following this research line, we propose a novel graph learning algorithm for RGB-T object tracking.

\subsubsection{End-to-End Deep Learning based Trackers}
Benefiting from deep learning techniques, many end-to-end deep trackers have been proposed and achieve excellent performance~\cite{nam2016mdnet,bertinetto2016fully,Cfnet2017cvpr,guo2017learning,he2018twofold}. Hyeonseob \emph{et al.}~\cite{nam2016mdnet} introduce the idea of multi-domain and train a domain-specific layers online for each video to adapt the target representations. It achieves outstanding performance on several benchmark datasets but has a long latency due to online training strategy. Bertinetto \emph{et al}. regard object tracking as similarity learning problem, and apply the Siamese network, named SiameFC, to visual tracking~\cite{bertinetto2016fully}. However, SiameFC lacks important online update strategy and cannot capture the change of appearance of objects. To handle this problem, Guo \emph{et al.}~\cite{guo2017learning} propose a dynamic Siamese network with a fast general transformation learning model. He \emph{et al.}~\cite{he2018twofold} design a twofold Siamese network that consists of a semantic branch and an appearance branch to improve the discrimination power of SiameFC in tracking. Besides, the correlation filter is also embedded into deep neural network~\cite{Cfnet2017cvpr}, named CFNet, to achieve end-to-end tracking.

\subsubsection{Correlation Filter based Trackers}
Correlation filters have obtained great achievement in visual tracking due to its accuracy and computational efficiency. The seminal work, MOSSE, is proposed by Bolme \emph{et al.}~\cite{bolme2010mosse}, which achieves hundreds of frames per second and high tracking accuracy. Recently, many advanced methods based on correlation filters are proposed in different aspects. For example, the CSK tracker~\cite{CSK12eccv} uses the kernel trick to improve performance. More and more discriminative features are utilized in tracking for robust representation of the objects, such as HOG~\cite{CSK15pami}, color names~\cite{danelljan2014adaptive} and deep features~\cite{ma2015hierarchical,qi2016hedged}. In addition, several trackers~\cite{danelljan2016beyond, danelljan2017eco} propose a continuous operation to integrate multiple features in a unified correlation filter framework. To adapt the size change, several adaptive scale processing trackers~\cite{ScaleCSK14bmvc,li2014scale} are investigated. To address the boundary effect, SRDCF~\cite{danelljan2015learning} is proposed by introducing spatial regularization into the learning of correlation filters to penalize the filter coefficients near the boundary. In CSR-DCF~\cite{lukezic2017csrdcf} tracker, spatial reliability map is constructed to adaptively select the object suitable for tracking.

\section{RGBT234 Tracking Benchmark}
This section will introduce the details of the newly created benchmark dataset, called RGBT234 in this paper, including imaging hardware setup, dataset annotation and statistics, baseline RGB and RGB-T trackers, and evaluation metrics.

\subsection{Dataset}
\label{sec::dataset}

\begin{figure}[!t]
	\includegraphics[width=1.0\columnwidth]{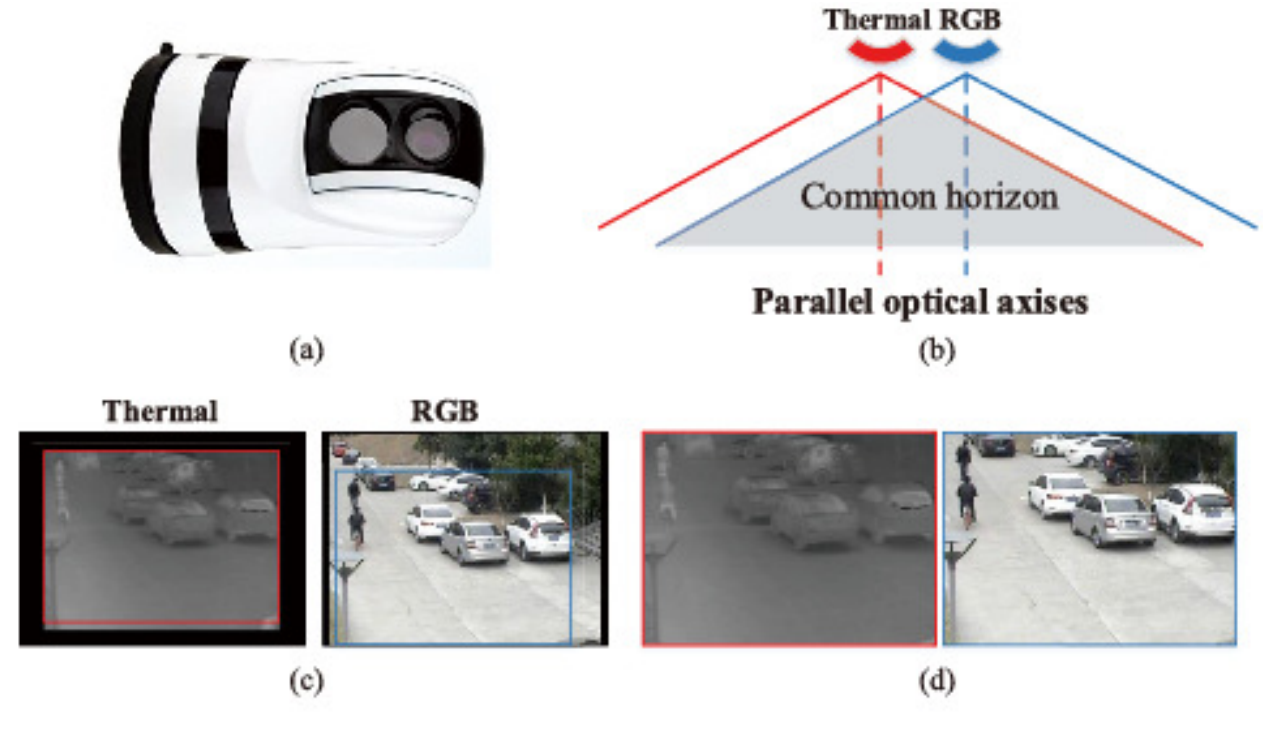}
	\caption{The mechanism of our imaging system for capturing RGB-T video sequences. (a) Illumination of the imaging hardware. (b) The optic axes of two cameras are aligned as parallel using the collimator. (c) Sample frame pairs captured by our system. The annotated bounding boxes indicate the common horizon, in which the colors denote the correspondences to the cameras. (d) Cropped image regions that are highly aligned. }
	\label{fig:ImagingPlatform}
\end{figure}

For large-scale performance evaluation of different RGB-T tracking algorithms, we collect 234 RGB-T videos, each containing a RGB video and a thermal video. In this subsection, we introduce the details of our imaging setup, dataset statistics and annotations.

\subsubsection{Imaging Setup}
Our imaging hardware consists of a turnable platform, a thermal infrared image (DLS-H37DM-A) and a CCD camera (SONY EXView HAD CC), as shown in Fig.~\ref{fig:ImagingPlatform} (a). In particular, these two cameras have same imaging parameters, and their optical axises are aligned as parallel by the collimator, as shown in Fig.~\ref{fig:ImagingPlatform} (b). The above setup makes the common horizon of these two cameras aligned in pixel level. Fig.~\ref{fig:ImagingPlatform} (c) and (d) show the details.


\subsubsection{Two-stage Annotation}
Different from our earlier work~\cite{Li17rgbt210}, we re-annotate our dataset more accurately in a two-stage fashion. In the first stage, we annotate each frame with a minimum bounding box covering the targets using more reliable source data. Second, we use the annotations in the first stage to adjust position of a minimum bounding box in less reliable source data. All annotations of two steps are done by a full-time annotator to guarantee consistency, respectively.

In addition, we also annotate the attributes for each video sequence for the attribute-sensitive performance analysis, as shown in Table~\ref{tab:2}. We present the attribute distribution over the entire dataset in Table~\ref{tab:1},  and some sample pairs with ground truths and attribute annotations in the project page:~\href{https://sites.google.com/view/ahutracking001/} {https://sites.google.com/view/ahutracking001/}. Note that each target object is annotated in every frame even when it is fully occluded.
\begin{table}[!t]
	\caption{List of attributes annotated in RGBT234}
	\begin{tabular}{ll}
		\hline\noalign{\smallskip}
		Attr & Description   \\
		\noalign{\smallskip}\hline\noalign{\smallskip}
		NO & No Occlusion - the target is not occluded.     \\
		PO & Partial Occlusion - the target object is partially occluded.       \\
		HO & \makecell[tl]{Heavy Occlusion - the target object is heavy occluded\\ (over 80$\%$ percentage).}     \\
		LI & \makecell[tl]{Low Illumination - the illumination in the target region \\is low.}      \\
		LR & Low Resolution - the resolution in the target region is low.       \\
		TC & \makecell[tl]{ Thermal Crossover - the target has similar temperature \\with other objects or background surroundings.}     \\
		DEF & Deformation - non-rigid object deformation.       \\
		FM & \makecell[tl]{Fast Motion - the motion of the ground truth between\\ two adjacent frames
			is larger than 20 pixels.}     \\
		SV & \makecell[tl]{Scale Variation - the ratio of the first bounding box \\and the current bounding box is out of the range [0.5,1].}\\
		MB & \makecell[tl]{Motion Blur - the target object motion results in the blur \\image
			information.}\\
		CM & \makecell[tl]{Camera Moving - the target object is captured by moving \\camera.}   \\
		BC & \makecell[tl]{Background Clutter - the background information which \\includes the target object is messy.}  \\
		\noalign{\smallskip}\hline
		\label{tab:2}       
	\end{tabular}
\end{table}

\begin{table*}\footnotesize 
	\caption{\footnotesize Distribution of visual attributes within RGBT234 dataset, showing the number of coincident attributes across all videos.}
	\centering
	\begin{tabular}{c c c c c c c c c c c c c}
		\hline
		& NO & PO & HO & LI & LR & TC & DEF & FM & SV & MB & CM & BC \\
		\hline
		NO & {\bf 41} & 0 & 0 & 8  & 10 & 10  & 12 & 7  & 26 & 9  & 16 & 9 \\
		PO & 0 &  {\bf 96} & 0 & 35  & 22 & 9  & 27 & 9  & 50 & 21  & 33 & 21 \\
		HO & 0 & 0 & {\bf 96} & 19  & 16 & 10  & 36 & 15  & 41 & 23  & 39 & 24\\
		LI & 8 & 35 & 19 & {\bf 63}  & 16 & 7  & 14 & 6  & 26 &  12 & 25 & 18 \\
		LR & 10 & 22 & 16 & 16  & {\bf 50} & 10  & 10 & 8  & 16 & 15  & 18 & 20 \\
		TC & 10 & 9 & 10 & 7  & 10 &  {\bf 28} & 9 &  7 & 11 & 10  & 10 & 7\\
		DEF & 12 & 27 & 36 & 14  & 10 &  9 & {\bf 76} & 16  & 36 & 12  & 33 & 20 \\
		FM & 7 & 9 & 15 & 6  & 8 & 7  & 16 & {\bf 32}  & 16 & 8  & 14 & 10 \\
		SV & 26 & 50 & 41 & 26  & 16 &  11 & 36 &  16 & {\bf 120} & 20  & 44 & 14\\
		MB & 9 & 21 & 23 & 12 & 15 & 10 & 12 & 8  & 20 & {\bf 55}  & 42 & 9 \\
		CM & 16 & 33 & 39 &  25 & 18 &  10 & 33 & 14  & 44 & 42  & {\bf 89} & 19 \\
		BC & 9 & 21 & 24 & 18 & 20 & 7 & 20 & 10 & 14 & 9 & 19 & {\bf 54} \\
		\hline
	\end{tabular}
	\label{tab:1}
\end{table*}
\subsubsection{Advantages over existing Datasets}
Table~\ref{tab:datasetComparation} provides summary of existing tracking datasets, including RGB, Thermal, and RGB-T ones. RGB datasets~\cite{RGBbenchmark13cvpr,RGBbenchmark15pami,vot14,vot15,liang2015Tcolor,Smeulders14pami,nus_pro2016} only provide RGB image sequences, which are sensitive to lighting conditions and could be easily affected by bad weathers, such as rain, smog and fog. These datasets are limited to ``good'' lighting and environmental conditions. Thermal datasets~\cite{OSUT2005,icra14ASL,cvpr14TIV,ICCV15TIR} only include thermal videos, and their weakness are revealed when thermal crossover occurs. RGB-T datasets~\cite{Davis07cviu,Torabi12cviu,Li16tip} can well address above issues, and the fusion of these two data enables long-term object tracking in day and night.

Our dataset is classified as a RGB-T dataset as it provides aligned RGB and thermal videos. Compared with other existing ones, our dataset has the following advantages: 1) it includes a large amount of annotated highly-accurate frames (total frames: about 234K, maximum frames per sequence: 8K), which allows trackers to perform large-scale performance evaluation. 2) Due to the superior imaging mechanism, its alignment across modalities is more accurate, and does not require any pre- and post-processing (\emph{e.g.}, stereo matching~\cite{Li16tip,KrotoskyT07} and color correction~\cite{Hwang_2015_CVPR}). 3) Its occlusion levels, including no, partial and heavy occlusions, are annotated for occlusion-sensitive evaluation of different algorithms. 4) Since the imaging parameters of RGB and thermal cameras in our platform are the same and their optical axis are parallel, its videos can be captured by both static and moving cameras while keeping the alignment accurate.
\subsection{Baseline Approaches}


To identify the importance of thermal data, we select several advanced RGB trackers for evaluations, including ECO~\cite{danelljan2017eco}, C-COT~\cite{danelljan2016beyond}, CFnet~\cite{Cfnet2017cvpr}, DSST~\cite{ScaleCSK14bmvc}, SOWP~\cite{Kim15iccv}, CSR-DCF~\cite{lukezic2017csrdcf}, SRDCF~\cite{danelljan2015learning} and SAMF~\cite{li2014scale}. The details of these trackers are presented in Table~\ref{tab:Evaluated_Methods}.

There are several RGB-T trackers, including L1-PF~\cite{Wu11icif}, JSR~\cite{Liu12infosci} and CSR~\cite{Li16tip}. We implement L1-PF and JSR according to their papers for comparison as these trackers do not release their code. In addition, we implement several new RGB-T baselines based on some RGB trackers for comprehensive evaluation, including SOWP~\cite{Kim15iccv}+RGBT(SVM-based tracker), MEEM~\cite{MEEM14eccv}+RGBT (SVM-based tracker), KCF~\cite{CSK15pami}+RGBT (correlation filter based tracker), CSR-DCF~\cite{lukezic2017csrdcf}+RGBT(correlation filter based tracker), CFnet~\cite{Cfnet2017cvpr}+RGBT(deep learning-based tracker). In particular, KCF+RGBT, SOWP+RGBT, MEEM+RGBT first concatenate RGB and thermal features into a single vector, and then employ the corresponding RGB trackers to perform object tracking. CSR-DCF+RGBT and CFnet+RGBT are implemented as follows: we first utilize the two-stream corresponding algorithms to track in different modalities data. Then the final tracking results are obtained by averaging tracking results in different modalities. We present the details of these RGB-T trackers in Table~\ref{tab:Evaluated_Methods}.
\begin{table*}[t]\scriptsize	
	\caption{\footnotesize Evaluated Tracking Algorithms.}
	\centering
	\begin{tabular}{l c c c c c c c c c}
		\hline\noalign{\smallskip}
		& \multicolumn{7}{c}{Representation}  &\multicolumn{2}{c}{Code}  \\
		\cline{2-8}
		\cline{9-10}
	    &\rotatebox{90}{Gray}& \rotatebox{90}{HOG} & \rotatebox{90}{Color Names} & \rotatebox{90}{Deep Feature} & \rotatebox{90}{Color Histogram} &\rotatebox{90}{CIE Lab} &\rotatebox{90}{Canny edge detector}& \rotatebox{90}{publish year} & \rotatebox{90}{RGB-T trackers}  \\\hline
    	L1-PF\cite{Wu11icif}& & &  &  & & &\checkmark &  2011 & \checkmark\\
		JSR\cite{Liu12infosci}& & &   &  & \checkmark&  & &2012 & \checkmark\\
		MEEM\cite{MEEM14eccv}+RGBT& & &    & & &\checkmark &  &2014 & \checkmark\\
		DSST\cite{ScaleCSK14bmvc}& &\checkmark &  & & & &  &2014 \\
		SAMF\cite{li2014scale}&\checkmark &\checkmark &  &  & & & & 2014\\
		KCF\cite{CSK15pami}+RGBT&\checkmark &\checkmark &  & &  & & & 2015 & \checkmark\\
		SRDCF\cite{danelljan2015learning}&\checkmark &\checkmark & \checkmark & & & & &2015\\
		SOWP\cite{Kim15iccv}& &\checkmark &  & & \checkmark & &  &2015  \\
		SOWP\cite{Kim15iccv}+RGBT& &\checkmark &  & & \checkmark & &  & 2015 &\checkmark  \\
		CSR\cite{Li16tip}& &\checkmark & & & &  & & 2016 & \checkmark\\
		C-COT\cite{danelljan2016beyond}& &\checkmark & \checkmark & \checkmark& & & & 2016  \\
		CSR-DCF\cite{lukezic2017csrdcf}&\checkmark &\checkmark & \checkmark &  & & & & 2017 \\	
		CSR-DCF\cite{lukezic2017csrdcf}+RGBT&\checkmark  &\checkmark & \checkmark &  & & & & 2017 &\checkmark\\	
		CFnet\cite{Cfnet2017cvpr}& & & &\checkmark &   & &  & 2017 &  \\	  	
	   	CFnet\cite{Cfnet2017cvpr}+RGBT& & & &\checkmark  &  & &  & 2017 & \checkmark \\	
	   	ECO\cite{danelljan2017eco}& &\checkmark  & \checkmark & \checkmark & & & & 2017  \\
	   	{\bf SGT}& &\checkmark & &   & \checkmark & &  &2017 & \checkmark \\
		\noalign{\smallskip}\hline	
	\end{tabular}
	\label{tab:Evaluated_Methods}
\end{table*}

\subsection{Evaluation Metrics}
There are 5 widely used metrics for tracking performance evaluation~\cite{RGBbenchmark15pami,vot15}, \emph{i.e.}, precision rate (PR), success rate (SR), Accuracy, Robustness, and expected average overlap (EAO). Similarly, we employ the following metrics for comprehensive evaluations.

\begin{itemize}
	\item {\bf Maximum Precision Rate (MPR)}. PR is the percentage of frames whose output location is within the given threshold distance of ground truth. That is to say, it computes the average Euclidean distance between the center locations of the tracked target and the manually labeled ground-truth positions of all the frames. Although our alignment between two modalities is highly accurate, there still exist small alignment errors. Therefore, we use maximum precision rate (MPR) instead of PR in this paper. Specifically, for each frame, we compute the above Euclidean distance on both RGB and thermal modalities, and adopt the smaller distance to compute the precision. According to~\cite{RGBbenchmark13cvpr,RGBbenchmark15pami}, we set the threshold to be 20 pixels to obtain the representative MPR.
	
	\item {\bf Maximum Success Rate (MSR).} SR is the ratio of the number of successful frames whose overlap is larger than a threshold. Given a tracked bounding box $ r_t $ and the ground-truth bounding box $ r_0 $ of a target object, the overlap score is defined as $ s= \frac{\mid r_t \cap r_0 \mid}{\mid r_t \cup r_0 \mid} $, where $ \cap $ and $ \cup $ represent the intersection and union operators, respectively, and $\mid \cdot \mid $ denotes the number of pixels in a region. Similar to MPR, we also define maximum success rate (MSR) to measure the tracker results. By varying the threshold, the MSR plot can be obtained, and we employ the area under curve of MSR plot to define the representative MSR.
	
	\item {\bf Accuracy.} Accuracy measures how well the predicted bounding box overlaps with the ground truth bounding box. It calculates the average overlap between the predicted and ground truth bounding boxes during successful tracking periods.
	
	\item {\bf Robustness.} Robustness measures how many times the tracker loses the target (fails) during tracking. A failure is indicated when the overlap measurement becomes zero. The per-frame accuracy is obtained as an average over these runs. Averaging per-frame accuracies gives per-sequence accuracy, while per-sequence robustness is computed by averaging failure rates over different runs.
	
	\item {\bf Expected Average Overlap (EAO).} Expected average overlap combines the raw values of per-frame accuracies and failures in a principled manner and has a clear practical interpretation~\cite{vot2016}. It averages the no-reset overlap of a tracker on several short-term sequences. In other words, it estimates how accurate the estimated bounding box is after a certain number of frames are processed since initialization~\cite{vot2016}.
	
\end{itemize}

\section{Proposed Approach}
\begin{figure}[!t]
	\includegraphics[width=0.8\columnwidth]{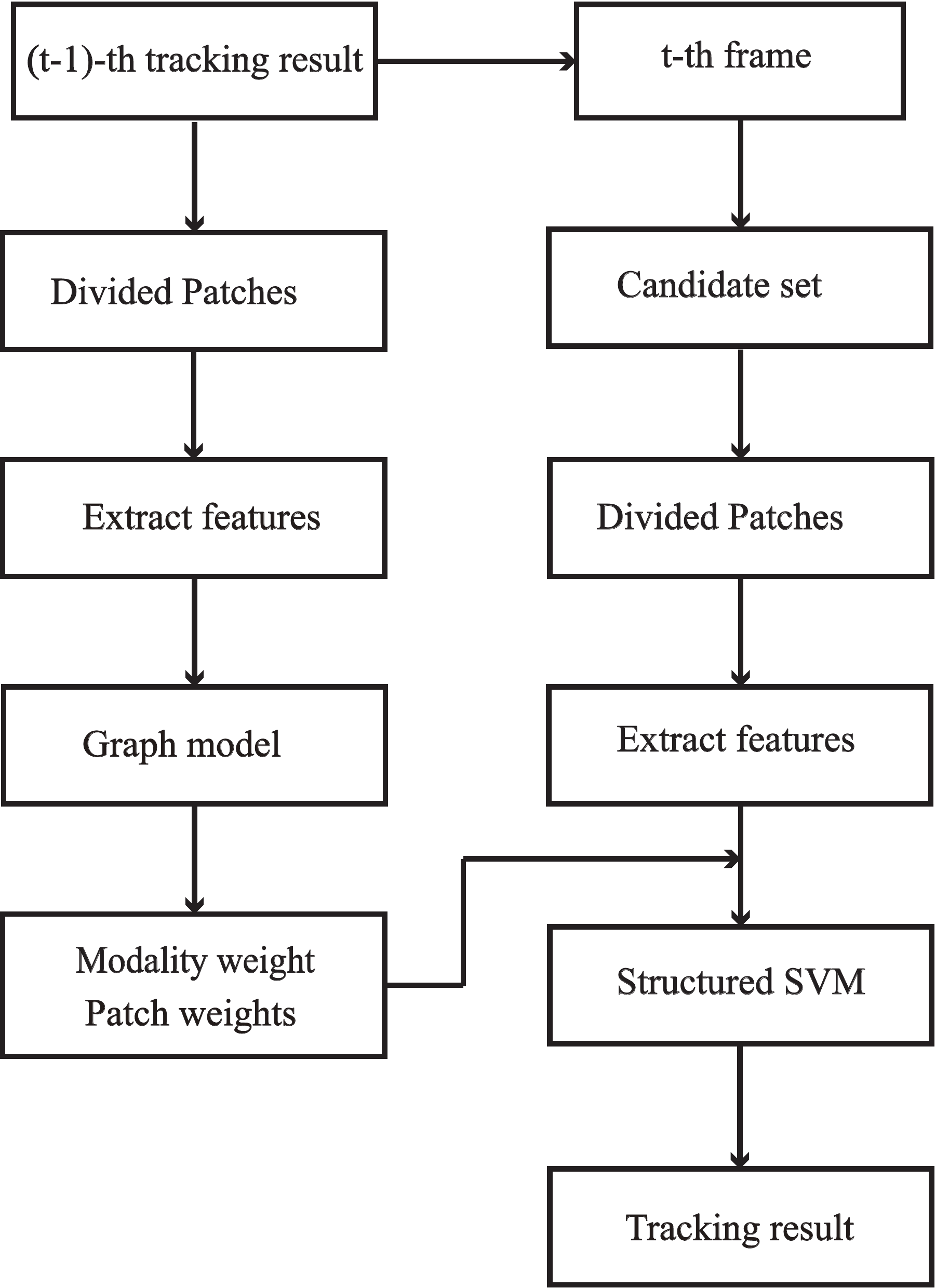}
	\centering
	\caption{Overflow of our tracking approach.}
	\label{fig:Framework}       
\end{figure}

Given the tracked bounding box, we partition it into non-overlapping patches, each of which is assigned with a weight to specify its importance to describe the target object. Moreover, we also associate the modality weights to different modalities to reflect their reliabilities. In particular, we propose a graph-based learning algorithm to compute the patch and modality weights together with graph structure optimization. Then, the patch and modality weights and patch features are combined into a single feature descriptor to represent the bounding box. The object tracking is finally performed via Structured SVM algorithm~\cite{Stuck11iccv}. It should be noted that other tracking-by-detection algorithms can also be used. Fig.~\ref{fig:Framework} shows the overflow of our tracking approach.

\subsection{Patch-based Representation Regularized Graph Learning}
\label{sec::WSRRGL}
In this subsection, we introduce the proposed graph-based learning algorithm that infers the patch weights and the modality weights.

\subsubsection{Formulation}
Each bounding box of the target object is partitioned into $n$ non-overlapping patches, and a set of low-level appearance features are extracted and further combined into a $d$-dimensional feature vector ${\bf x}^m_i$ for characterizing the $i$-th patch in the $m$-th modality. All the feature descriptors of $n$ patches in one bounding box form a data matrix ${\bf X}^m=\{{\bf x}^m_1,{\bf x}^m_2,...,{\bf x}^m_n\}\in {\mathbb R}^{d\times n}$, where $m$ indicates the index of the modality with the range between 1 and $M$. Herein, we discuss the general case for the scalability, and RGB-T data used in this paper is the special case with $M=2$.

We take all the above patches as graph nodes and dynamically learn a graph according to the intrinsic relationship of patches instead of using fixed-structure graph in conventional methods~\cite{Zhou04nips,Kim15iccv}. Motivated by the sparse clustering algorithms~\cite{SSC2009,SSLSR2009}, we assume the foreground or background patches are in the same sparse subspace~\cite{SSC2009}, and thus each patch descriptor can be sparsely self-represented by a linear combination of remaining patches: ${\bf X}^m={\bf X}^m{\bf Z}^m$, where ${\bf Z}^m \in {\mathbb R}^{n\times n}$ is a sparse representation coefficient matrix. Sparse constraints can automatically select most informative neighbors for each patch (higher-order relationship), making the graph more powerful and discriminative~\cite{SSLSR2009}. Considering the patches are often disturbed by noises and/or corruptions, we introduce a noise matrix to improve the robustness. The joint sparse representation with the convex relaxation~\cite{SSLSR2009} for all modalities can be formulated as:
\begin{equation}\label{Eq::sparse_representaion}
\begin{split}
& \min_{\bf Z,E}\sum^{\rm M}_{m=1}\|{\bf X}^m-{\bf X}^m{\bf Z}^m-{\bf E}^m\|^2_F+\lambda\|{\bf E}^m\|_{2,1}+\gamma\|{\bf Z}\|_{2,1},
\end{split}
\end{equation}
where $||\cdot||_F$ and $||\cdot||_{2,1}$ denote the Frobenius norm and the $l_{2,1}$ norm of a matrix, respectively. $\lambda$ and $\gamma$ are the balanced parameters. ${\bf Z}=[{\bf Z}^1;...;{\bf Z}^M]\in {\mathbb R}^{Mn\times n}$ is the joint sparse representation coefficients matrix, and $\|{\bf Z}\|_{2,1}$ encourages each patch shares the same pattern across different modalities. ${\bf E}^m\in {\mathbb R}^{d\times n}$ denotes the noise matrix, and $\|{\bf E}^m\|_{2,1}$ makes it as the sparse sample-specific corruptions, \emph{i.e.}, some patches are corrupted and others are clean. It is worth noting that other
constraints on the representation matrix, such as low rank~\cite{LiuLYSYM13},
can also be incorporated in~\eqref{Eq::sparse_representaion}, and we only employ the sparse constraints in this paper with an emphasis on efficient performance.

In~\eqref{Eq::sparse_representaion}, different modalities contribute equally, but usually have different imaging qualities in real-life scenarios. Therefore, we assign a weight to each modality to represent the reliability to deal with occasional perturbation or malfunction of individual source, which allows our method to integrate different spectrum data adaptively~\cite{Li16tip}. We integrate these modality weights in~\eqref{Eq::sparse_representaion}, and have
\begin{equation}\label{Eq::weighted_sparse_representation}
\begin{split}
& \min_{\bf Z,E,r} \sum^{M}_{m=1}(\frac{(r^m)^2}{2}\|{\bf X}^m-{\bf X}^m{\bf Z}^m-{\bf E}^m \|^2_F+\lambda\|{\bf E}^m\|_{2,1})\\
&+ \gamma\|{\bf Z}\|_{2,1} + \Gamma\|{\bf 1-r} \|^2_F,
\end{split}
\end{equation}
where ${\bf r}=[r^1,...,r^{M}]^T$ is the modality weighting vector. $r^m$ is the modality weight in the $m$-th modality. In general, the reconstruction error can measure how well the patch could be sparsely reconstructed from other patches. Therefore, the qualities of different modalities can be reflected by their respective reconstruction errors. From the first term in~\eqref{Eq::weighted_sparse_representation}, we can see that our method places larger weights on those modalities which have smaller reconstruction errors, resulting in a quality-aware weight optimization. The last term of~\eqref{Eq::weighted_sparse_representation} is the regularization of $r^m$, which avoids a degenerate solution of $r^m$ while allowing them to be specified independently. $\Gamma$ is an adaptive parameter, which is determined after the first iteration~\cite{Li17rgbt210}.

\begin{figure}[!t]
	\centering
	\includegraphics[width=\columnwidth]{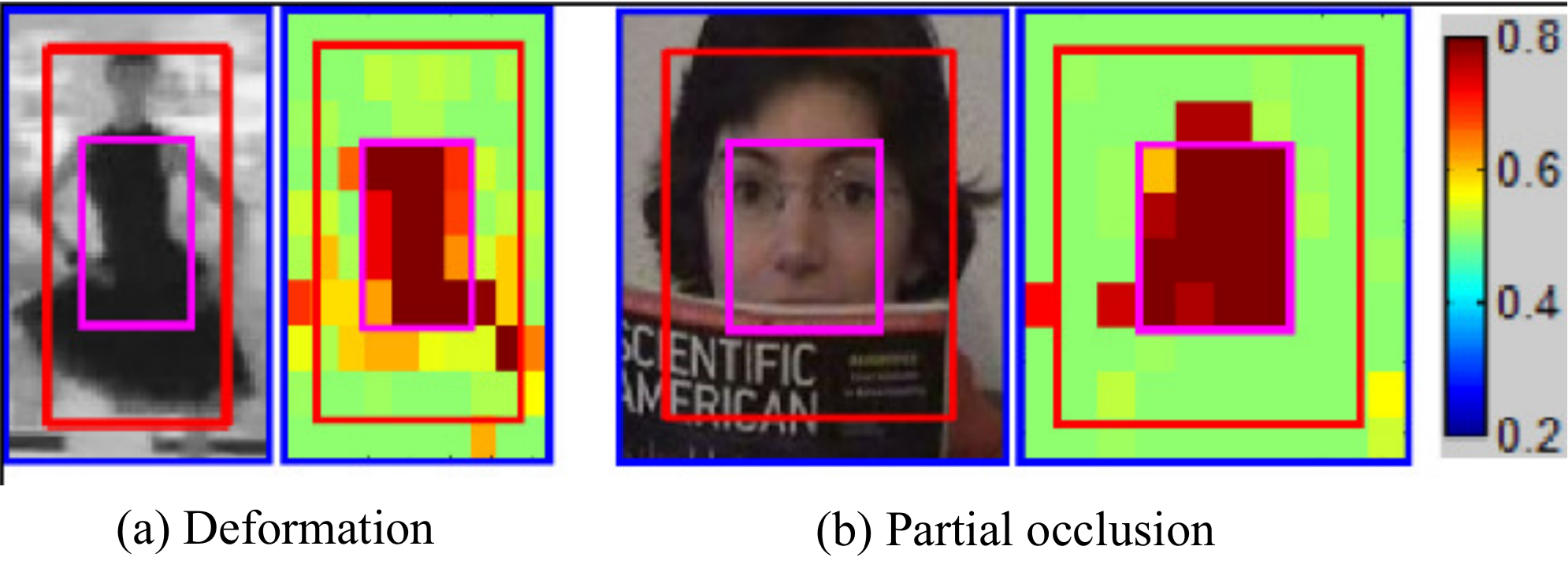}
	\caption{Two samples for showing the optimized patch weights are shown in (a) and (b). The original, shrunk and expanded bounding boxes are represented by the red, pink and blue colors, respectively. The optimized patch weights are also shown for clarity, in which the hotter color indicates the larger weight.}\label{fig::weight_samples}
\end{figure}

Most of methods utilize the representation coefficient matrix to define the graph affinity by $\frac{|{\bf Z}|+|{\bf Z}^T|}{2}$. Although this kind of affinity is somehow valid, its meaning is already different from the original definition~\cite{ijcai/Guo15a}. Recall that we assume the patch descriptor should have a larger probability to be in the same cluster if their representations have a smaller distance. Therefore, instead of directly using sparse representations, we learn a more meaningful graph affinity ${\bf A}$ by the following constraints:
\begin{equation}\label{Eq::graph_affinity}
\begin{split}
& \min_{\bf A} \delta \sum^n_{i,j=1}\|{\bf Z}_i-{\bf Z}_j\|^2_F{\bf A}_{ij}+ \frac{\omega}{2}\|{\bf A}\|^2_F,\\
& s.t. {\bf A}^T{\bf 1}= {\bf 1},{\bf A}\geq0,
\end{split}
\end{equation}
where $\delta$, $\omega$ are the balanced parameters, and ${\bf 1}$ denotes a unit vector. ${\bf A}\in {\mathbb R}^{n\times n}$ is the desired affinity matrix. ${\bf A}_{ij}$ reflects the probability of the patch $i$ and $j$ from the same class based on the distance between their joint sparse representations ${\bf Z}_i$ and ${\bf Z}_j$ across all modalities. The constraints ${\bf A}^T{\bf 1}= {\bf 1}$ and ${\bf A}\geq0$ are to guarantee the probability property of ${\bf A}_i$. The last term is to avoid overfitting of ${\bf A}$.

Given the graph affinity ${\bf A}$, we can compute the patch node weights in a semi-supervized way. Let ${\bf q}=\{q_1,q_2,...,q_n\}^T$ be an initial weight vector, in which $q_i=1$ if $q_i$ is a foreground patch, and $q_i=0$ is a background patch. ${\bf q}$ is computed by the initial ground-truth (for first frame) or tracking results (for subsequent frames) as follows: for the $i$-th patch, $q_i=1$ if it belongs to the shrunk region of bounding box and the remaining patches are 0. Fig.~\ref{fig::weight_samples} shows the details. Similar to the PageRank and spectral clustering algorithm~\cite{nips02sc}, the patch weights ${\bf s}$ can be calculated as follows:
\begin{equation}\label{Eq::ranking}
\begin{split}
& \min_{\bf s} \alpha\sum^n_{i,j=1}({s}_i-{s}_j)^2{\bf A}_{ij}+\beta\|{\bf s-q} \|^2_F,
\end{split}
\end{equation}
where $\alpha$ and $\beta$ are the balanced parameters. The first term is the smoothness constraint and the second term is the fitting constraint.

In this paper, we aim to jointly optimize the modality weights, the sparse representations, and the graph (including the structure, the edge weights and the node weights) for boosting their respective performance. Therefore, the final formulation of the proposed model can be written by combining~\eqref{Eq::weighted_sparse_representation},~\eqref{Eq::graph_affinity} and~\eqref{Eq::ranking}:
\begin{equation}
\label{Eq::ObjectModel}
\begin{split}
& \min_{{\bf Z, E, r,s,A}} \sum^{M}_{m=1}(\frac{(r^m)^2}{2}\|{\bf X}^m-{\bf X}^m{\bf Z}^m-{\bf E}^m \|^2_F+\lambda\|{\bf E}^m\|_{2,1})\\
&+ \gamma\|{\bf Z}\|_{2,1} + \delta\sum^n_{i,j=1}\|{\bf Z}_i-{\bf Z}_j \|^2_F{\bf A}_{ij} + \alpha\sum^n_{i,j=1}({s}_i-{s}_j)^2{\bf A}_{ij}\\
& + \beta\|{\bf s-q} \|^2_F + \Gamma\|{\bf 1-r} \|^2_F + \frac{\omega}{2}\|{\bf A}\|^2_F,\\
& s.t. {\bf A}^T{\bf 1}= {\bf 1},{\bf A}\geq0,
\end{split}
\end{equation}

\subsubsection{Optimization}
\begin{figure}[!t]
	\includegraphics[width=0.6\columnwidth]{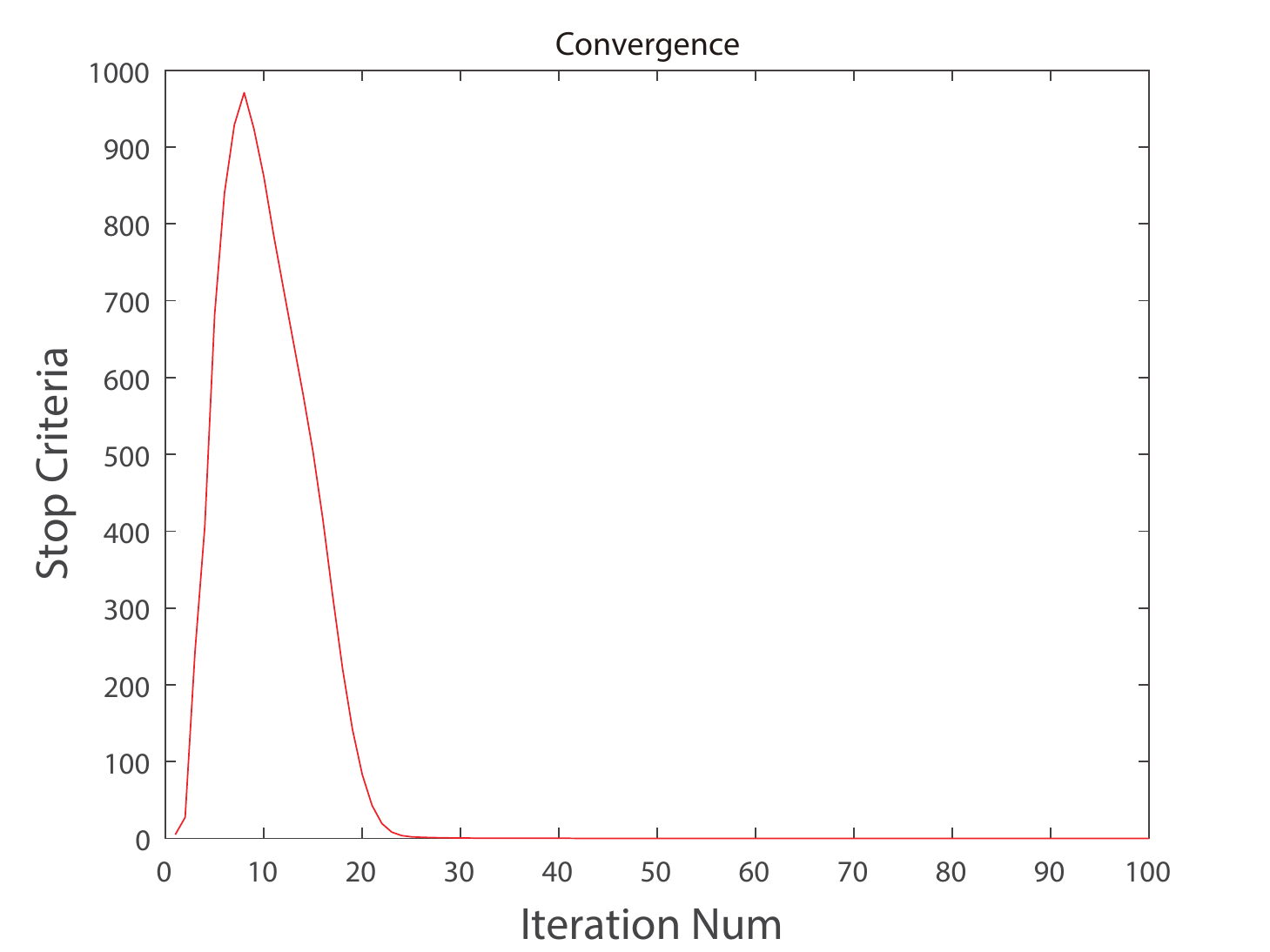}
	\centering
	\caption{Convergence of the proposed model on all video sequences of RGBT234.}
	\label{fig:Convergence}       
\end{figure}

We design an efficient ADMM (alternating direction method of multipliers) algorithm~\cite{NIPS2011ADM} to optimize~\eqref{Eq::ObjectModel}, and the efficiency is demonstrated in Fig.~\ref{fig:Convergence}. Although our model is non-convex, the subproblem of each variable with fixing others is convex and has a closed form solution. Since each subproblem is convex, we can guarantee that the limit point by our algorithm satisfies the Nash equilibrium conditions~\cite{SIAM2013Nash}. The detailed optimization procedure can be referred to our previous work~\cite{Li17rgbt210}.

Although~\eqref{Eq::ObjectModel} seems complex, its parameters are easy to adjust, and the tracking performance is insensitive to parameter variations~\cite{Li17rgbt210}. The final weight of the $i$-th patch is computed by $\hat{s}_i={1}/{(1+\exp{(-\sigma s_i)})}$, where the parameter $\sigma$ is fixed to be 37 in this work.

\subsection{Tracking via Structured SVM }
\subsubsection{Feature Representation}
In RGB-T tracking, it is important to construct a robust descriptor
$\Psi(\hat{\bf x},y)$ of the bounding box $y$. Therefore, we assign a weight $\hat s_i$ to each patch to suppress the background effects, and also assign a weight $r^m$ for each modality to integrate different source data adaptively. Combining these weights with the corresponding patch feature descriptor ${\bf x}^m_i$ forms the final feature representation of the bounding box:
\begin{equation}
\label{Eq:feature_representation}
\begin{aligned}
&\Psi({\bf \hat{x}}_t,y_t)=[r^1_{t-1}\hat{s}_{t-1,1}{\bf x}^1_{t,1},..., r^1_{t-1}\hat{s}_{t-1,n}{\bf x}^1_{t,n},\\
&...,r^M_{t-1}\hat{s}_{t-1,1}{\bf x}^M_{t,1},..., r^M_{t-1}\hat{s}_{t-1,n}{\bf x}^M_{t,n}]^T.
\end{aligned}
\end{equation}

From~\eqref{Eq:feature_representation}, we can see that $\hat s_i$ and $r^m$ play a critical role in the target feature representation.

\subsubsection{Tracking}
We adopt the structured SVM (S-SVM)~\cite{Stuck11iccv} to perform object tracking in this paper. Note that scale variations emerge frequently during tracking, and thus translation estimation only cannot provide accurate results. In other words, the search space excludes the true state of a target regarding scale. The problem can be alleviated by considering candidates of various sizes. However, this approach may generate too many candidates, increase false positives, and reduce the tracking reliability. Therefore, we decompose the problem of target state estimation into two subproblems: translation estimation and scale estimation~\cite{danelljan2014accurate,Ma15cvpr} as follows.

\subsubsection{Translation estimation}
In this work, we combine object representations of multiple modalities with S-SVM to achieve robust RGB-T tracking. Given the bounding box of the target object in previous frame $t-1$, we first set a searching window in current frame $t$, and sample a set of candidates within the searching window. S-SVM selects the optimal target bounding box $y^*_t$ in the t-th frame by maximizing a classifier score:
\begin{equation}
\label {eq:SSVM}
y^*_t=\arg\max_{y_t}~(\nu{{\bf h}_{t-1}^T\Psi(y_t)}+(1-\nu){\bf h}_{0}^T\Psi(\hat{\bf x}_t,y_t)),
\end{equation}
where $\pi$ is a balancing parameter, and ${\bf h}_{t-1}$ is the normal vector of a decision plane of $(t-1)$-th frame. Let $\Psi(y_t)$ denote the object descriptor representing a bounding box $y_t$ at the $t$-th frame, and ${\bf h}_{0}$ is learned in the initial frame, which can prevent it from learning drastic appearance changes~\cite{pami/MatthewsIB04}. Instead of using binary-labeled samples, S-SVM employs structured sample that consists of a target bounding box and nearby boxes in the same frame to prevent the labelling ambiguity in training the classifier. Specifically, it constraints that the confidence score of a target bounding box $y_t$ is larger than that of nearby box $y$ by a margin determined by the intersection over union overlap ratio (denoted as $IoU(y_t,y)$) between two boxes:
\begin{equation}
\label{eq:decision-plane}
{\bf h}^*=\arg\min_{\bf h}~\xi||{\bf h}||^2+\sum_{\bf y}\max\{0,\triangle(y_t,y)-{\bf h}^T\epsilon(y_t,y)\},
\end{equation}
where $\triangle(y_t,y)=1-IoU(y_t,y)$, $\epsilon(y_t,y)=\Psi(x_t,y_t)-\Psi(x,y)$, and $\xi=0.0001$ is a regularization parameter. By this way, S-SVM can reduce adverse effects of false labelling. To prevent the effects of unreliable tracking results, we update the classifier only when the confidence score of tracking result is larger than a threshold $\theta$. The confidence score of tracking result in $t$-th frame is defined as the average similarity between the weighted descriptor of the tracked bounding box and the positive support vectors: ${\frac{1}{|\mathbb S_{t}|}}\sum_{{\bf s}\in{\mathbb S}_{t}}{\bf s}^T\Psi(y_t^*)$, where $\mathbb{S}_{t}$ is the set of the positive support vectors at time $t$.

\subsubsection{Scale estimation}
During RGB-T tracking, we construct a target pyramid around the estimated translation location for scale estimation~\cite{Ma15cvpr}. Let $W\times H$ be the target size in a test frame and N indicates the number of scales: $\mathbb{B}=\{a^{\bar{n}}|\bar{n}=\lfloor-\frac{N-1}{2}\rceil,\lfloor-\frac{N-3}{2}\rceil,...,\lfloor\frac{N-1}{2}\rceil\}$. For each $ b\in B $, we extract an image region of size $ bW \times bH $ centered around the estimated location and obtain its classification score $ y^*_{t,b} $. Then, we uniformly resize all image regions with the size $W\times H$, and the optimal scale $ b $ of target can be estimated by maximizing
the classification scores of all resized image regions:
\begin{equation}
b^*_t=\arg\max_{b_t\in\mathbb{B}}(\pi{{\bf h}_{t-1}^T\Psi(y_{t,b})}+(1-\pi){\bf h}_{0}^T\Psi(\hat{\bf x}_{t,b},y_{t,b})).
\end{equation}

\section{Performance Evaluation}
In this section, we first present the evaluation details, and then show experimental results of 14 tracking algorithms on the RGBT234 dataset with detailed analysis of each challenging factor. At last, the limitations of the proposed approach are analyzed in detail. The source code, the new dataset, the baseline trackers and the evaluation results will be available online for free academic usage and accessible reproducible research.

\begin{figure*}[!t]
	\includegraphics[width=1.6\columnwidth]{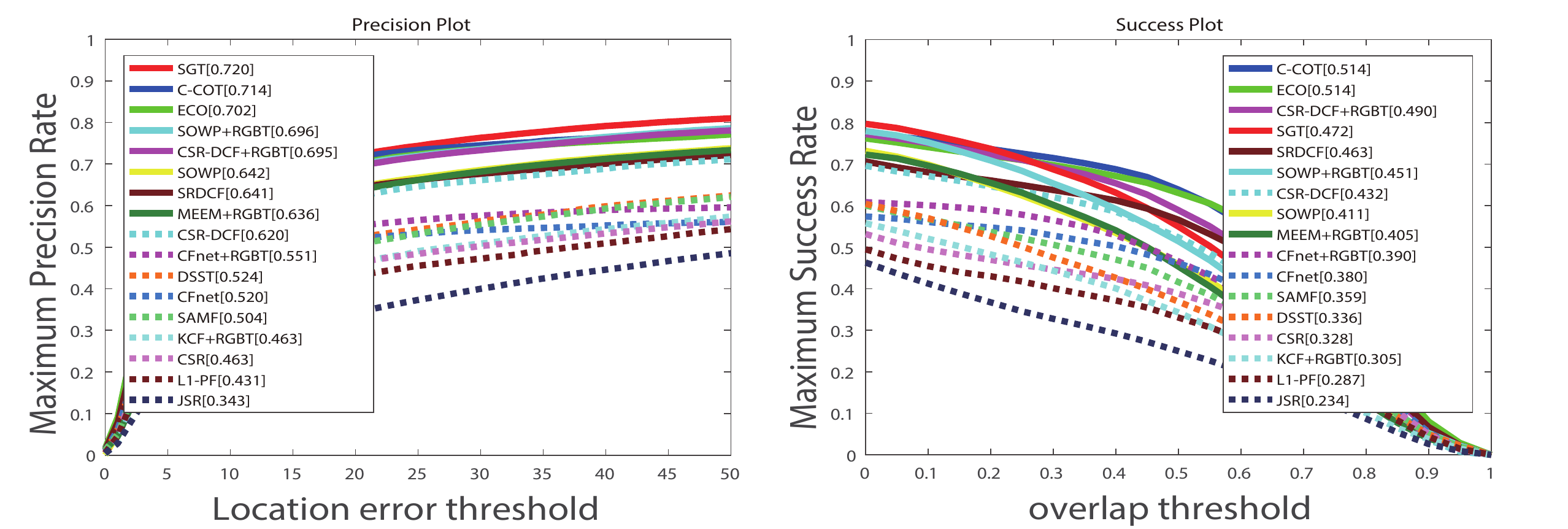}
    \centering
	\caption{MPR and MSR plots of different tracking algorithms on RGBT234 dataset.}
	\label{fig:ALL_results}       
\end{figure*}

\subsection{Parameter Settings}
For fair comparison, we fix all parameters and other settings of the proposed approach in all experiments. We partition the bounding box into 64 non-overlapping patches to balance accuracy and efficiency, and extract color (gray for thermal source) and gradient histograms for each patch. The dimension of gradient and each color channel is set to be 8. To improve efficiency, each frame is scaled so that the minimum side length of bounding box is 32 pixels, and the side length of a searching window is fixed to be $2\sqrt{\bar{W}\bar{H}}$, where $\bar{W}$ and $\bar{H}$ are the width and height of the scaled bounding box, respectively.
	
Although~\eqref{Eq::ObjectModel} seems very complex, its parameters are easy to tune. On one hand, similar to ~\cite{guo2015robust}, we simplify the setting as $\lambda=\gamma$. Following~\cite{guo2015robust}, we set $\{\lambda,\gamma,\delta,\omega\}=\{0.1,0.1,11,1\}$. On the other hand, $\alpha$ and $\beta$ control the balance of smoothness and fitness of ${\bf s}$. According to the setting of similar models~\cite{yang2013saliency,li2017learning}, we set $\{\alpha,\beta\}=\{10,0.15\}$. In the structured SVM, we empirically set $\{\nu,\zeta\}=\{0.67,0.35\}~$\cite{Kim15iccv}. All parameters are optimized by varying them within a certain scope.

\subsection{Overall Performance}
The overall performance of all the trackers is summarized by the maximum success and precision plots as shown in Fig.~\ref{fig:ALL_results}. Table~\ref{tb::Accurary-Robustness} presents the evaluation results using other 3 metrics, \emph{i.e.}, Accuracy, Robustness, and Expected Average Overlap (EAO). We analyze these tracking algorithms from the following two lines.

\subsubsection{Tracking Methods Without Deep Features}
The proposed approach (\emph{i.e.}, SGT) performs best over others in MPR and EAO (combination of Accuracy and Robustness), but slightly worse than CSR-DCF~\cite{lukezic2017csrdcf}+RGBT in MSR (by 1.8\%). The lower performance in MSR and Accuracy against others suggests that SGT cannot handle scale variations well, and we will study robust scale handling algorithms in the future. Overall, the results demonstrate the effectiveness of SGT in learning RGB-T object features for visual tracking. Also, we have the following major observations and conclusions. For RGB-T methods, SGT, SOWP~\cite{Kim15iccv}+RGBT and CSR-DCF+RGBT rank top 3 and significantly outperform others. Note that SGT and SOWP+RGBT are to learn weighted patch representations and CSR-DCF+RGBT is to integrate segmentation into correlation filter learning, which show the effectiveness of background suppression in RGB-T tracking. Similar observations can also be discovered in the performance comparison of RGB trackers. SGT outperforms SOWP+RGBT suggesting the effectiveness of adaptive fusion of RGB and thermal data. By observing the comparison results between RGB and RGB-T methods, the great performance shows the importance of thermal information in visual tracking.

\begin{table}[t]\scriptsize 
	\caption{\footnotesize The following table represents the Accuracy (A), Robustness (R), and Expected Average Overlap (EAO) of these evaluated trackers. The top one is marked with red color, the second highest is marked with green color, and the third highest is marked with blue color}
	\centering		
		\begin{tabular}{ c | c c c }
			\hline			
			& Accuracy & Robustness & EAO \\\hline		
			L1-PF & 0.58 & 4.87 & 0.1514 \\\hline		
			JSR & 0.52 &10.88 & 0.0767 \\\hline		
			MEEM+RGBT & 0.55 &2.05 & 0.2500  \\\hline		
			DSST & 0.17 &3.68 & 0.0492  \\\hline
			SAMF & 0.61 & 3.39 & 0.1938 \\\hline
			KCF+RGBT & 0.55 & 3.88 &0.1782 \\\hline	
			SRDCF & \textcolor{blue}{0.65} &2.34 & 0.2465 \\\hline
			SOWP & 0.57 &2.10 & 0.2673\\\hline
			SOWP+RGBT  & 0.57 & \textcolor{blue}{1.66} &\textcolor{blue}{0.2950} \\\hline
			CSR & 0.21 &10.31 &0.0280 \\\hline
			C-COT & \textcolor{green}{0.66}& \textcolor{red}{1.63} &\textcolor{green}{0.3019} \\\hline
			CSR-DCF & 0.63 &2.10 & 0.2755\\\hline
			CSR-DCF+RGBT & 0.63 & 1.75 & 0.2938 \\\hline
			CFnet & 0.62 & 2.71 &0.2275 \\\hline
			CFnet+RGBT & 0.60 & 2.59 &0.2220 \\\hline	
			ECO & \textcolor{red}{0.67} &1.76  & 0.2937 \\\hline
			SGT & 0.59 &\textcolor{green}{1.63} & \textcolor{red}{0.3076} \\\hline
	\end{tabular}
	\label{tb::Accurary-Robustness}
\end{table}

\subsubsection{Tracking Methods Based on Deep Features}
For comprehensive evaluation, we also select several state-of-the-art RGB trackers using deep features in our platform, including ECO~\cite{danelljan2017eco}, C-COT~\cite{danelljan2016beyond}, CFnet~\cite{Cfnet2017cvpr} and CFnet+RGBT. Benefit from thermal data, SGT outperforms them in MPR and EAO (by 1.8\%/0.0139, 0.6\%/0.0057, 20.0\%/0.0801 and 16.9\%/0.0856 in MPR/EAO for ECO, C-COT, CFnet and CFnet+RGBT, respectively). Due to weak scale handling scheme, SGT achieves lower performance than ECO and C-COT in MSR and Accuracy, and achieves lower performance than C-COT in Robustness. Note that SGT has the following advantages over the deep learning based methods. 1) It does not rely on large-scale annotated training samples. We train our model using the ground truth in the first frame, and update it in subsequent frames. 2) It is easy to implement as each subproblem of the proposed model has a closed-form solution.

\begin{table*}[t]\scriptsize 
	\caption{\footnotesize Attribute-based Maximum Precision Rate and Maximum Success Rate (MPR/MSR \%) of different tracking algorithms on RGBT234 dataset, including ECO~\cite{danelljan2017eco}, C-COT~\cite{danelljan2016beyond}, CFnet~\cite{Cfnet2017cvpr}, CSR-DCF~\cite{lukezic2017csrdcf}, SRDCF~\cite{danelljan2015learning}, SAMF~\cite{li2014scale}, DSST~\cite{danelljan2014accurate}, SOWP~\cite{Kim15iccv}, KCF~\cite{CSK15pami}, MEEM~\cite{MEEM14eccv}, CSR~\cite{Li16tip}, JSR~\cite{Liu12infosci} and L1-PF~\cite{Wu11icif}. The best, second and third best performance are marked in \textcolor{red}{red}, \textcolor{green}{green} and \textcolor{blue}{blue} colors, respectively.}
	\centering
	\resizebox{\textwidth}{15mm}{		
		\begin{tabular}{ c | c c c c c c c c c c c c c c c c c}
			\hline			
			& L1-PF & JSR & MEEM+RGBT & DSST &SAMF & KCF+RGBT & SRDCF & SOWP & SOWP+RGBT & CSR & C-COT & CSR-DCF & CSR-DCF+RGBT & CFnet  & CFnet+RGBT  & ECO & SGT  \\\hline
			NO & 56.5/37.9 & 56.7/41.5 & 74.1/47.4 & 69.7/43.3 & 67.6/48.5 & 57.1/37.1  & 79.1/\textcolor{blue}{58.5} & 80.1/50.2  & 86.8/53.7 &56.7/41.5 & \textcolor{red}{88.8}/\textcolor{red}{65.6}  & 78.8/56.6  &82.6/\textcolor{blue}{60.0} & 72.4/54.5  & 76.4/56.3 & \textcolor{green}{88.0}/\textcolor{green}{65.5} & \textcolor{blue}{87.7}/55.5 \\
			PO & 47.5/31.4 & 37.8/25.2 & 68.3/42.9 & 56.5/36.2 & 54.0/38.0  & 52.6/34.4 &68.8/49.9   & 66.6/42.7  & \textcolor{green}{74.7}/48.4 & 49.4/34.9 &\textcolor{blue}{74.1}/\textcolor{red}{54.1} & 64.1/44.4  & 73.7/52.2 & 57.7/41.8 & 59.7/41.7  &72.2/\textcolor{green}{53.4} &\textcolor{red}{77.9}/51.3 \\
			HO & 33.2/22.2 & 25.9/18.1 & 54.0/34.9 & 41.0/27.0 & 39.8/28.5 & 35.6/23.9& 52.6/37.1 & 54.7/35.4 & 57.0/37.9 & 38.4/26.8 &\textcolor{red}{60.9}/\textcolor{green}{42.7} & 52.2/36.0 & \textcolor{blue}{59.3}/\textcolor{blue}{40.9} & 37.9/27.2 & 41.7/29.0 & \textcolor{green}{60.4}/\textcolor{red}{43.2} & 59.2/39.4 \\
			LI & 40.1/26.0 &38.1/26.4 & 67.1/42.1 & 48.3/29.9 &46.8/32.7 & 51.8/34.0 & 57.5/40.9  & 52.4/33.6 & \textcolor{red}{72.3}/\textcolor{green}{46.8} & 39.3/27.3 & 64.8/45.4 & 49.0/32.9 &\textcolor{blue}{69.1}/\textcolor{red}{47.4} & 43.6/31.5 & 52.3/36.9  & 63.5/45.0 & \textcolor{green}{70.5}/\textcolor{blue}{46.2} \\
			LR & 46.9/27.4 &39.9/23.9& 60.8/37.3 & 57.9/36.8 &50.7/32.9 & 49.2/31.3& 62.0/41.0 & 67.9/42.1 & \textcolor{blue}{72.5}/46.2 & 41.3/25.9 & \textcolor{green}{73.1}/\textcolor{red}{49.4}&57.9/37.0 & 72.0/\textcolor{green}{47.6} & 48.2/33.9  & 55.1/36.5  & 68.7/\textcolor{blue}{46.4} & \textcolor{red}{75.1}/\textcolor{green}{47.6} \\
			TC & 37.5/23.8 & 33.8//20.9 & 61.2/40.8 & 49.5/32.5 & 54.7/38.1  & 38.7/25.0  & 66.1/46.7 &71.2/46.2 & 70.1/44.2 & 44.4/32.5 &\textcolor{red}{84.0}/\textcolor{red}{61.0}  & 62.9/42.8 &66.8/46.2&51.2/38.5  & 45.7/32.7  &\textcolor{green}{82.1}/\textcolor{green}{60.9} & \textcolor{blue}{76.0}/\textcolor{blue}{47.0} \\
			DEF & 36.4/24.4 &27.8/20.2& 61.7/41.3 & 43.8/32.5 & 42.4/33.2 & 41.0/29.6  & 56.3/40.6 & 61.1/42.0 & \textcolor{green}{65.0}/46.0 & 44.8/33.1 & \textcolor{blue}{63.4}/\textcolor{green}{46.3} &55.7/41.0 & 63.0/\textcolor{blue}{46.2}  &46.0/34.0  & 52.3/36.7 & 62.2/45.8 & \textcolor{red}{68.5}/\textcolor{red}{47.4} \\
			FM & 32.0/19.6 &25.6/15.7 & 59.7/36.5 & 35.5/22.4 & 42.4/27.0 & 37.9/22.3& 52.6/34.3  & 57.9/33.5 & \textcolor{green}{63.7}/38.7 & 34.9/22.0 & \textcolor{blue}{62.8}/\textcolor{red}{41.8} & 53.0/35.0 & 52.9/35.8 & 36.3/25.3  &37.6/25.0  & 57.0/\textcolor{blue}{39.5} &\textcolor{red}{67.7}/\textcolor{green}{40.2} \\
			SV & 45.5/30.6 &35.5/23.8 & 61.6/37.6 & 56.8/33.7 & 55.4/39.7 & 44.1/28.7 & 70.4/\textcolor{blue}{51.8} & 66.6/39.6 &66.4/40.4 & 50.9/37.3 & \textcolor{red}{76.2}/\textcolor{red}{56.2} & 67.0/47.3 & \textcolor{blue}{70.7}/49.9 & 59.5/43.2  & 59.6/43.3  & \textcolor{green}{74.0}/\textcolor{green}{55.8} & 69.2/43.4 \\
			MB & 28.6/20.6 &24.2/17.2 &55.1/36.7 & 35.8/25.1 & 37.8/27.9 & 32.3/22.1 & 55.9/41.5 & 59.8/39.9 & 63.9/42.1 & 37.9/27.0 & \textcolor{green}{67.3}/\textcolor{green}{49.5} & 55.0/39.8 & 58.0/42.5 & 38.4/29.4  & 35.7/27.1  & \textcolor{red}{68.9}/\textcolor{red}{52.3} & \textcolor{blue}{64.7}/\textcolor{blue}{43.6} \\
			CM & 31.6/22.5 &29.1/21.0 &58.5/38.3 & 39.9/27.9 & 40.2/30.6 & 40.1/27.8 & 56.9/40.9 & 59.8/39.0 & \textcolor{blue}{65.2}/43.0 & 41.5/30.1 &  \textcolor{green}{65.9}/ \textcolor{green}{47.3}& 55.8/39.6  & 61.1/44.5 & 41.7/32.1 &41.7/31.8 &63.9/\textcolor{red}{47.7} & \textcolor{red}{66.7}/\textcolor{blue}{45.2} \\
			BC & 34.2/22.0 & 33.2/21.2 & \textcolor{blue}{62.9}/38.3 & 45.8/29.3 & 37.6/25.9 & 42.9/27.5 & 48.1/32.4 & 52.8/33.6 & \textcolor{green}{64.7}/\textcolor{red}{41.9} & 38.8/25.3 & 59.1/39.9 & 50.3/32.4 & 61.8/\textcolor{blue}{41.0} & 36.3/25.7 & 46.3/30.8  & 57.9/39.9 &\textcolor{red}{65.8}/\textcolor{green}{41.8} \\\hline
			ALL  & 43.1/28.7 &34.3/23.4 & 63.6/40.5 & 52.4/33.6 & 50.4/35.9 & 46.3/30.5 & 64.1/46.3  & 64.2/41.1 &69.6/45.1& 46.3/32.8 & \textcolor{green}{71.4}/\textcolor{red}{51.4} & 62.0/43.2 &69.5/\textcolor{green}{49.0} & 52.0/38.0  &55.1/39.0 & \textcolor{blue}{70.2}/\textcolor{red}{51.4} & \textcolor{red}{72.0}/\textcolor{blue}{47.2} \\\hline
	\end{tabular}}
	\label{tb::AttributeResults}
\end{table*}

\subsection{Attribute-based Performance}
To quantify the advantages of different tracking algorithms on RGB234, we present the attribute-based MPR and MSR scores in Table~\ref{tb::AttributeResults}, and analyze the details in the following paragraphs.

\subsubsection{Occlusion}
For partial occlusion (PO), SGT and SOWP+RGBT perform best in MPR and comparable against ECO and C-COT, which suggest that PO introduces some background information that could easily corrupt object detectors or trackers, but the weighted local patch representation is an effective way to alleviate background effects. When heavy occlusion (HO) occurs, their performance decreases much more than ECO and C-COT as the initial seeds in SGT and SOWP+RGBT are unreliable which might result in inaccurate weights in background suppression.

\subsubsection{Low Illumination}
Low illumination (LI) indicates that the imaging quality of visible spectrum is bad, and the thermal information is thus critical for robust tracking. From Table~\ref{tb::AttributeResults}, we can see that SGT, SOWP+RGBT and CSR-DCF+RGBT significantly outperform ECO, C-COT, SOWP, CSR-DCF and others. It demonstrates the importance of thermal data in visual tracking.
	
\subsubsection{Low Resolution}
Although deep learning is a powerful tool in feature representation, it usually performs not well for objects with very low resolution (LR). As shown in Table~\ref{tb::AttributeResults}, SGT achieves superior performance over C-COT and ECO in MPR. Therefore, learning non-deep features like in SGT might be a better direction for  low-resolution objects tracking.	

\subsubsection{Thermal Crossover}
C-COT, ECO and SGT are ranked the top 3 on thermal crossover (TC) and significantly excel the remaining methods. It suggests these trackers could learn a good feature representation to discriminate objects from background in the TC challenge. It is worth noting that SOWP achieves better results than SOWP+RGBT on TC. It suggests that directly fusing RGB and thermal information sometimes is ineffective as some individual source might be noisy data.

\subsubsection{Deformation}
The targets with deformation (DEF) usually have irregular shapes and their shapes also vary over time. In such circumstance, the bounding box cannot represent the target well and background suppression is essential for avoiding trackers or detectors corrupted. It is verified by the big superior performance of SGT over other methods and the comparable results of SOWP, SOWP+RGBT and CSR-DCF+RGBT against ECO and C-COT.
	
\subsubsection{Fast Motion}
The performance on the fast motion (FM) subset depends on the search strategies of trackers. For example, correlation filter-based methods pad original target bounding box with a factor, which is usually small to balance accuracy-efficiency trade-off, while structured SVM-based trackers often set a searching window in a more flexible way. The results verify it, such as the significant performance gain of SOWP+RGBT over CSR-DCF+RGBT and the superior performance of SGT over ECO and C-COT. In addition, the comparable performance of CSR-DCF and CSR-DCF+RGBT, CFnet and CFnet+RGBT also supports the conclusion in TC that the direct fusion strategy sometimes is not a good way for RGB-T tracking.

\subsubsection{Scale Variation}
From the results, we can see that correlation filter- and deep learning-based methods can handle scale variations well due to the discriminative ability of them in slightly different appearances, such as ECO, C-COT, CSR-DCF+RGBT and SRDCF. The performance of CFnet and CFnet+RGBT, SOWP and SOWP+RGBT are comparable, which is similar to TC and FM.

\subsubsection{Motion Blur}
In general, motion blur (MB) results in appearance degradation. And how to learn a good feature is critical to address the MB challenge. Deep learning-based trackers and weighted patch representation-based trackers (like ECO, C-COT, SGT and SOWP-RGBT) excel others in MPR and MSR. Some tracking methods, which regard thermal as an extra channel, may not perform effective integration of different data (for example, CFnet+RGBT is much worse than CFnet).

\subsubsection{Camera Moving}
Camera moving (CM) is also a challenging factor for visual tracking, which usually introduces motion variation and motion blur. From the results, we find SGT can adapt to CM well, followed by C-COT, ECO and SOWP-RGBT. On CM, CFnet+RGBT does not bring much performance gain over CFnet.

\subsubsection{Background Clutter}
For background clutter (BC), SGT and SOWP+RGBT outperform other trackers with a clear margin. It suggests the weighted patch representations and thermal information are beneficial to mitigating background effects in the bounding box descriptions.

In summary, the following major conclusions can be made by the RGBT234 evaluation. 1) Thermal data plays a critical role in visual tracking. 2) SGT could learn a good feature representation via the graph learning algorithm for RGB-T tracking.  3) The weighted patch feature representations and deep learning features are two potential ways for RGB-T tracking performance boosting. 3) Utilizing non-deep features (like in SGT) might be a better direction for low-resolution objects tracking. 4) Direct integration of RGB and thermal information sometimes is ineffective in RGB-T tracking.

\section{Concluding Remark}
In this paper, we carry out large scale experiments to evaluate the performance of recent online tracking algorithms, including RGB-, RGB-T- and deep learning-based ones. The results demonstrate that by incorporating depth data, trackers can achieve better performance and handle a variety of challenges much more reliable. We hope that our unified benchmark provides new insights to the field, by making experimental evaluation more standardized and easily accessible.

From the evaluation results, trackers that utilize thermal have advantages especially when the target is small, deformable or under partial occlusion, low illumination, camera moving and background clutter. Target appearance might be disturbed significantly under deformation, occlusion, low illumination and camera moving. And low resolution and background clutter make detection difficult, which are the main causes of model drifting for traditional RGB trackers. However, thermal data can provide complementary information to RGB ones to improve appearance features under deformation, occlusion, camera moving and low resolution, and is also insensitive to lighting conditions and thus performs well in low illumination and background clutter. On the contrary, visible spectrum information can discriminate targets from background when thermal data are unreliable, \emph{e.g.}, thermal crossover occurs.

In addition, based on our evaluation results and observations, we highlight some tracking components which are essential for RGB-T tracking performance improvement. First, adaptive fusion is effective for RGB-T tracking. Modality weights that identify modality qualities can be incorporated into tracking models or approaches to achieve this goal (\emph{e.g.}, SGT). Second, background suppression is critical for effective RGB-T tracking. It can be achieved by weighting local features (\emph{e.g.}, SGT, SOWP+RGBT) and integrating segmentation mask in model learning (\emph{e.g.}, CSR-DCF+RGBT). Third, powerful feature representations are essential for high-performance RGB-T tracking. It can be achieved by incorporating deep learning features (\emph{e.g.}, ECO and C-COT) or weighted features (SGT and SOWP+RGBT) into tracking framework. Finally, components used in conventional RGB trackers are also important for RGB-T tracking, such as motion model especially when the motion of target is large or abrupt. And local models are particularly useful when the appearance of target is partially changed, such as partial occlusion or deformation. Improving above mentioned components will further advance the state-of-the-art of RGB-T object tracking.

\ifCLASSOPTIONcaptionsoff
  \newpage
\fi



\bibliographystyle{IEEEtran}
\bibliography{mybibfile}

\begin{thebibliography}{10}
\providecommand{\url}[1]{#1}
\csname url@samestyle\endcsname
\providecommand{\newblock}{\relax}
\providecommand{\bibinfo}[2]{#2}
\providecommand{\BIBentrySTDinterwordspacing}{\spaceskip=0pt\relax}
\providecommand{\BIBentryALTinterwordstretchfactor}{4}
\providecommand{\BIBentryALTinterwordspacing}{\spaceskip=\fontdimen2\font plus
\BIBentryALTinterwordstretchfactor\fontdimen3\font minus
  \fontdimen4\font\relax}
\providecommand{\BIBforeignlanguage}[2]{{%
\expandafter\ifx\csname l@#1\endcsname\relax
\typeout{** WARNING: IEEEtran.bst: No hyphenation pattern has been}%
\typeout{** loaded for the language `#1'. Using the pattern for}%
\typeout{** the default language instead.}%
\else
\language=\csname l@#1\endcsname
\fi
#2}}
\providecommand{\BIBdecl}{\relax}
\BIBdecl

\bibitem{ThermalApplications14mva}
R.~Gade and T.~B. Moeslund, ``Thermal cameras and applications: a survey,''
  \emph{Machine Vision and Applications}, vol.~25, no.~1, pp. 245--262, 2014.

\bibitem{Liu12infosci}
H.~Liu and F.~Sun, ``Fusion tracking in color and infrared images using joint
  sparse representation,'' \emph{Information Sciences}, vol.~55, no.~3, pp.
  590--599, 2012.

\bibitem{WELD16tcsvt}
C.~Li, X.~Wang, L.~Zhang, J.~Tang, H.~Wu, and L.~Lin, ``Weld: Weighted low-rank
  decomposition for robust grayscale-thermal foreground detection,'' \emph{IEEE
  Transactions on Circuits and Systems for Video Technology}, vol.~27, no.~4,
  pp. 725--738, 2016.

\bibitem{Davis07cviu}
J.~W. Davis and V.~Sharma, ``Background-subtraction using contour-based fusion
  of thermal and visible imagery,'' \emph{Computer Vision and Image
  Understanding}, vol. 106, no.~2, pp. 162--182, 2007.

\bibitem{Torabi12cviu}
A.~Torabi, G.~Masse, and G.-A. Bilodeau, ``An iterative integrated framework
  for thermal-visible image registration, sensor fusion, and people tracking
  for video surveillance applications,'' \emph{Computer Vision and Image
  Understanding}, vol. 116, no.~2, pp. 210--221, 2012.

\bibitem{Li16tip}
C.~Li, H.~Cheng, S.~Hu, X.~Liu, J.~Tang, and L.~Lin, ``Learning collaborative
  sparse representation for grayscale-thermal tracking,'' \emph{IEEE
  Transactions on Image Processing}, vol.~25, no.~12, pp. 5743--5756, 2016.

\bibitem{KrotoskyT07}
S.~J. Krotosky and M.~M. Trivedi, ``On color-, infrared-, and multimodal-stereo
  approaches to pedestrian detection,'' \emph{{IEEE} Trans. Intelligent
  Transportation Systems}, vol.~8, no.~4, pp. 619--629, 2007.

\bibitem{Hwang_2015_CVPR}
S.~Hwang, J.~Park, N.~Kim, and et~al., ``Multispectral pedestrian detection:
  Benchmark dataset and baseline,'' in \emph{The IEEE Conference on Computer
  Vision and Pattern Recognition (CVPR)}, 2015.

\bibitem{Tsochantaridis05jmlr}
I.~Tsochantaridis, T.~Joachims, T.~Hofmann, and Y.~Altun, ``Large margin
  methods for structured and interdependent output variables,'' \emph{Journal
  of Machine Learning Research}, vol.~6, pp. 1453--1484, 2005.

\bibitem{Zhou04nips}
D.~Zhou, J.~Weston, A.~Gretton, O.~Bousquet, and B.~Scholkopf, ``Ranking on
  data manifolds,'' in \emph{Proceedings of Neural Information Processing
  Systems}, 2004.

\bibitem{Kim15iccv}
H.-U. Kim, D.-Y. Lee, J.-Y. Sim, and C.-S. Kim, ``Sowp: Spatially ordered and
  weighted patch descriptor for visual tracking,'' in \emph{Proceedings of IEEE
  International Conference on Computer Vision}, 2015.

\bibitem{Li17aaai}
C.~Li, L.~Lin, W.~Zuo, and J.~Tang, ``Learning patch-based dynamic graph for
  visual tracking,'' in \emph{Proceedings of the Thirty-First {AAAI} Conference
  on Artificial Intelligence}, 2017, pp. 4126--4132.

\bibitem{ijcai/Guo15a}
X.~Guo, ``Robust subspace segmentation by simultaneously learning data
  representations and their affinity matrix,'' in \emph{Proceedings of the
  Twenty-Fourth International Joint Conference on Artificial Intelligence},
  2015.

\bibitem{Lan14cvpr}
X.~Lan, A.~J. Ma, and P.~C. Yuen, ``Multi-cue visual tracking using robust
  feature-level fusion based on joint sparse representation,'' in
  \emph{Proceedings of IEEE Conference on Computer Vision and Pattern
  Recognition}, 2014.

\bibitem{LiuLYSYM13}
G.~Liu, Z.~Lin, S.~Yan, J.~Sun, Y.~Yu, and Y.~Ma, ``Robust recovery of subspace
  structures by low-rank representation,'' \emph{IEEE Transactions on Pattern
  Analysis and Machine Intelligence}, vol.~35, no.~1, pp. 171--184, 2013.

\bibitem{NIPS2011ADM}
Z.~Lin, R.~Liu, and Z.~Su, ``Linearized alternating direction method with
  adaptive penalty for low-rank representation,'' in \emph{Advances in Neural
  Information Processing Systems 24}, 2011.

\bibitem{RGBbenchmark13cvpr}
Y.~Wu, J.~Lim, and M.-H. Yang, ``Online object tracking: A benchmark,'' in
  \emph{Proceedings of IEEE Conference on Computer Vision and Pattern
  Recognition}, 2013.

\bibitem{RGBbenchmark15pami}
------, ``Object tracking benchmark,'' \emph{IEEE Transactions on Pattern
  Analysis and Machine Intelligence}, vol.~37, no.~9, pp. 1834--1848, 2015.

\bibitem{vot14}
M.~Kristan, R.~Pflugfelder, A.~Leonardis, J.~Matas, and L.~C. et~al., ``The
  visual object tracking vot2014 challenge results,'' in \emph{Proceedings in
  European Conference on Computer Vision}, 2014.

\bibitem{vot15}
M.~Kristan, J.~Matas, A.~Leonardis, and M.~F. et~al., ``The visual object
  tracking vot2015 challenge results,'' in \emph{ICCV workshop on Visual Object
  Tracking Challenge}, 2015.

\bibitem{liang2015Tcolor}
P.~Liang, E.~Blasch, and H.~Ling, ``Encoding color information for visual
  tracking: Algorithms and benchmark,'' \emph{IEEE Transactions on Image
  Processing}, vol.~24, no.~12, pp. 5630--5644, 2015.

\bibitem{Smeulders14pami}
A.~W.~M. Smeulders, D.~M. Chu, R.~Cucchiara, S.~Calderara, A.~Dehghan, and
  M.~Shah, ``Visual tracking: An experimental survey,'' \emph{IEEE Transactions
  on Pattern Analysis and Machine Intelligence}, vol.~36, no.~7, pp.
  1442--1468, 2014.

\bibitem{nus_pro2016}
A.~Li, M.~Lin, Y.~Wu, M.~Yang, and S.~Yan, ``{NUS-PRO: A New Visual Tracking
  Challenge},'' \emph{IEEE Transactions on Pattern Analysis and Machine
  Intelligence}, vol.~38, no.~2, pp. 335--349, 2016.

\bibitem{OSUT2005}
J.~W. Davis and M.~A. Keck, ``A two-stage template approach to person detection
  in thermal imagery,'' in \emph{Application of Computer Vision, 2005.
  WACV/MOTIONS '05 Volume 1. Seventh IEEE Workshops on}, 2005.

\bibitem{icra14ASL}
J.~Portmann, S.~Lynen, M.~Chli, and R.~Siegwart, ``People detection and
  tracking from aerial thermal views,'' in \emph{Proceedings of IEEE
  International Conference on Robotics and Automation (ICRA)}, 2014.

\bibitem{cvpr14TIV}
Z.~Wu, N.~Fuller, D.~Theriault, and M.~Betke, ``A thermal infrared video
  benchmark for visual analysis,'' in \emph{2014 IEEE Conference on Computer
  Vision and Pattern Recognition Workshops}, 2014.

\bibitem{ICCV15TIR}
M.~Felsberg, A.~Berg, G.~Hager, and e.~a. Ahlberg, ``The thermal infrared
  visual object tracking vot-tir2015 challenge results,'' in \emph{The IEEE
  International Conference on Computer Vision (ICCV) Workshops}, 2015.

\bibitem{Li17rgbt210}
C.~Li, N.~Zhao, Y.~Lu, C.~Zhu, and J.~Tang, ``Weighted sparse representation
  regularized graph learning for rgb-t object tracking,'' in \emph{Proceedings
  of ACM International Conference on Multimedia}, 2017.

\bibitem{Bilodeau14ipt}
G.-A. Bilodeau, A.~Torabi, and P.-L. S.-C. et~al., ``Thermal-visible
  registration of human silhouettes: A similarity measure performance
  evaluation,'' \emph{Infrared Physics \& Technology}, vol.~64, pp. 79--86,
  2014.

\bibitem{Cvejic07cvpr}
N.~Cvejic, S.~G. Nikolov, H.~D. Knowles, A.~Loza, A.~Achim, D.~R. Bull, and
  C.~N. Canagarajah, ``The effect of pixel-level fusion on object tracking in
  multi-sensor surveillance video,'' in \emph{Proceedings of IEEE Conference on
  Computer Vision and Pattern Recognition}, 2007.

\bibitem{Wu11icif}
Y.~Wu, E.~Blasch, G.~Chen, L.~Bai, and H.~Ling, ``Multiple source data fusion
  via sparse representation for robust visual tracking,'' in \emph{Proceedings
  of International Conference on Information Fusion}, 2011.

\bibitem{Li17tsmcs}
C.~Li, X.~Sun, X.~Wang, L.~Zhang, and J.~Tang, ``Grayscale-thermal object
  tracking via multi-task laplacian sparse representation,'' \emph{IEEE
  Transactions on Systems, Man, and Cybernetics: Systems}, vol.~47, no.~4, pp.
  673--681, 2017.

\bibitem{LMM2005Struck}
I.~Tsochantaridis, T.~Joachims, T.~Hofmann, and Y.~Altun, ``Large margin
  methods for structured and interdependent output variables,'' \emph{J. Mach.
  Learn. Res.}, vol.~6, pp. 1453--1484, 2005.

\bibitem{Stuck11iccv}
S.~Hare, A.~Saffari, and P.~H.~S. Torr, ``Struck: Structured output tracking
  with kernels,'' in \emph{Proceedings of IEEE International Conference on
  Computer Vision}, 2011.

\bibitem{hare2016struck}
S.~Hare, S.~Golodetz, A.~Saffari, V.~Vineet, M.-M. Cheng, S.~L. Hicks, and
  P.~H. Torr, ``Struck: Structured output tracking with kernels,'' \emph{IEEE
  Transactions on Pattern Analysis and Machine Intelligence}, vol.~38, no.~10,
  pp. 2096--2109, 2016.

\bibitem{Li17regle}
C.~Li, X.~Wu, Z.~Bao, and J.~Tang, ``Regle: Spatially regularzied graph
  learning for visual tracking,'' in \emph{Proceedings of ACM International
  Conference on Multimedia}, 2017.

\bibitem{nam2016mdnet}
H.~Nam and B.~Han, ``Learning multi-domain convolutional neural networks for
  visual tracking,'' in \emph{Computer Vision and Pattern Recognition (CVPR),
  2016 IEEE Conference on}, 2016, pp. 4293--4302.

\bibitem{bertinetto2016fully}
L.~Bertinetto, J.~Valmadre, J.~F. Henriques, A.~Vedaldi, and P.~H. Torr,
  ``Fully-convolutional siamese networks for object tracking,'' in
  \emph{European conference on computer vision}, 2016, pp. 850--865.

\bibitem{Cfnet2017cvpr}
J.~Valmadre, L.~Bertinetto, J.~Henriques, A.~Vedaldi, and P.~H. Torr,
  ``End-to-end representation learning for correlation filter based tracking,''
  in \emph{Computer Vision and Pattern Recognition (CVPR), 2017 IEEE Conference
  on}, 2017, pp. 5000--5008.

\bibitem{guo2017learning}
Q.~Guo, W.~Feng, C.~Zhou, R.~Huang, L.~Wan, and S.~Wang, ``Learning dynamic
  siamese network for visual object tracking,'' in \emph{Proceedings of the
  IEEE International Conference on Computer Vision}, 2017, pp. 1--9.

\bibitem{he2018twofold}
A.~He, C.~Luo, X.~Tian, and W.~Zeng, ``A twofold siamese network for real-time
  object tracking,'' \emph{arXiv preprint arXiv:1802.08817}, 2018.

\bibitem{bolme2010mosse}
D.~S. Bolme, J.~R. Beveridge, B.~A. Draper, and Y.~M. Lui, ``Visual object
  tracking using adaptive correlation filters,'' in \emph{Computer Vision and
  Pattern Recognition (CVPR), 2010 IEEE Conference on}, 2010, pp. 2544--2550.

\bibitem{CSK12eccv}
J.~F. Henriques, R.~Caseiro, P.~Martins, and J.~Batista, ``Exploiting the
  circulant structure of tracking-by-detection with kernels,'' in
  \emph{Proceedings of European Conference on Computer Vision}, 2012.

\bibitem{CSK15pami}
------, ``High-speed tracking with kernelized correlation filters,'' \emph{IEEE
  Transactions on Pattern Analysis and Machine Intelligence}, 2015.

\bibitem{danelljan2014adaptive}
M.~Danelljan, F.~Shahbaz~Khan, M.~Felsberg, and J.~Van~de Weijer, ``Adaptive
  color attributes for real-time visual tracking,'' in \emph{IEEE Conference on
  Computer Vision and Pattern Recognition (CVPR)}, 2014, pp. 1090--1097.

\bibitem{ma2015hierarchical}
C.~Ma, J.-B. Huang, X.~Yang, and M.-H. Yang, ``Hierarchical convolutional
  features for visual tracking,'' in \emph{Proceedings of the IEEE
  International Conference on Computer Vision}, 2015, pp. 3074--3082.

\bibitem{qi2016hedged}
Y.~Qi, S.~Zhang, L.~Qin, H.~Yao, Q.~Huang, J.~Lim, and M.-H. Yang, ``Hedged
  deep tracking,'' in \emph{Proceedings of the IEEE Conference on Computer
  Vision and Pattern Recognition}, 2016, pp. 4303--4311.

\bibitem{danelljan2016beyond}
M.~Danelljan, A.~Robinson, F.~S. Khan, and M.~Felsberg, ``Beyond correlation
  filters: Learning continuous convolution operators for visual tracking,'' in
  \emph{European Conference on Computer Vision}, 2016.

\bibitem{danelljan2017eco}
M.~Danelljan, G.~Bhat, F.~S. Khan, and M.~Felsberg, ``Eco: Efficient
  convolution operators for tracking,'' in \emph{Proceedings of the 2017 IEEE
  Conference on Computer Vision and Pattern Recognition (CVPR)}, 2017.

\bibitem{ScaleCSK14bmvc}
M.~Danelljan, G.~Hager, F.~Khan, and M.~Felsberg, ``Accurate scale estimation
  for robust visual tracking,'' in \emph{Proceedings of British Machine Vision
  Conference}, 2014.

\bibitem{li2014scale}
Y.~Li and J.~Zhu, ``A scale adaptive kernel correlation filter tracker with
  feature integration,'' in \emph{European Conference on Computer Vision},
  2014.

\bibitem{danelljan2015learning}
M.~Danelljan, G.~Hager, F.~Shahbaz~Khan, and M.~Felsberg, ``Learning spatially
  regularized correlation filters for visual tracking,'' in \emph{Proceedings
  of the IEEE International Conference on Computer Vision}, 2015.

\bibitem{lukezic2017csrdcf}
A.~Lukezic, T.~Vojir, L.~Cehovin, J.~Matas, and M.~Kristan, ``Discriminative
  correlation filter with channel and spatial reliability,'' in
  \emph{Proceedings of the IEEE Conference on Computer Vision and Pattern
  Recognition}, 2017.

\bibitem{MEEM14eccv}
J.~Zhang, S.~Ma, and S.~Sclaroff, ``{MEEM:} robust tracking via multiple
  experts using entropy minimization,'' in \emph{Proceedings of European
  Conference on Computer Vision}, 2014.

\bibitem{vot2016}
M.~K. et~al., ``The visual object tracking vot2016 challenge results,'' in
  \emph{Proceedings in European Conference on Computer Vision (ECCV)
  workshops}, 2016.

\bibitem{SSC2009}
E.~Elhamifar and R.~Vidal, ``Sparse subspace clustering,'' in \emph{Proceedings
  of IEEE Conference on Computer Vision and Pattern Recognition}, 2009.

\bibitem{SSLSR2009}
S.~Yan and H.~Wang, ``Semi-supervised learning by sparse representation,'' in
  \emph{Proceedings of the SIAM International Conference on Data Mining}, 2009.

\bibitem{nips02sc}
A.~Y. Ng, M.~I. Jordan, and Y.~Weiss, ``On spectral clustering: Analysis and an
  algorithm,'' in \emph{Proceedings of Advances in Neural Information
  Processing Systems}, 2001.

\bibitem{SIAM2013Nash}
Y.~Xu and W.~Yin, ``A block coordinate descent method for regularized
  multiconvex optimization with applications to nonnegative tensor
  factorization and completion,'' \emph{SIAM Journal on Imaging Sciences},
  vol.~6, no.~3, pp. 1758--1789, 2013.

\bibitem{danelljan2014accurate}
M.~Danelljan, G.~H{\"a}ger, F.~Khan, and M.~Felsberg, ``Accurate scale
  estimation for robust visual tracking,'' in \emph{Proceedings of British
  Machine Vision Conference}, 2014.

\bibitem{Ma15cvpr}
C.~Ma, X.~Yang, C.~Zhang, and M.-H. Yang, ``Long-term correlation tracking,''
  in \emph{Proceedings of IEEE Conference on Computer Vision and Pattern
  Recognition}, 2015.

\bibitem{pami/MatthewsIB04}
I.~A. Matthews, T.~Ishikawa, and S.~Baker, ``The template update problem,''
  \emph{IEEE Transactions on Pattern Analysis and Machine Intelligence},
  vol.~26, no.~6, pp. 810--815, 2004.

\bibitem{guo2015robust}
X.~Guo, ``Robust subspace segmentation by simultaneously learning data
  representations and their affinity matrix.'' in \emph{IJCAI}, 2015.

\bibitem{yang2013saliency}
C.~Yang, L.~Zhang, H.~Lu, X.~Ruan, and M.-H. Yang, ``Saliency detection via
  graph-based manifold ranking,'' in \emph{Computer Vision and Pattern
  Recognition (CVPR), 2013 IEEE Conference on}, 2013.

\bibitem{li2017learning}
C.~Li, L.~Lin, W.~Zuo, and J.~Tang, ``Learning patch-based dynamic graph for
  visual tracking.'' in \emph{AAAI}, 2017.

\end{thebibliography}

\end{document}